\documentclass[journal]{IEEEtran}
\usepackage{cite}
\usepackage[colorlinks,
linkcolor=black,
anchorcolor=black,
citecolor=black,
urlcolor=black
]{hyperref}
\usepackage[english]{babel}
\usepackage{graphicx}
\graphicspath{{Figures/}}
\DeclareGraphicsExtensions{.pdf,.jpeg,.png}
\usepackage{amsmath,amssymb}
\usepackage{gensymb}
\usepackage{xcolor,colortbl,arydshln}
\usepackage{lipsum}
\usepackage[mathlines,switch]{lineno}
\usepackage[ruled, boxed, linesnumbered]{algorithm2e}
\usepackage{algpseudocode}
\usepackage{multirow}
\usepackage{colortbl}% http://ctan.org/pkg/colortbl
\usepackage{xcolor}% http://ctan.org/pkg/xcolor
\usepackage{mathrsfs}
\usepackage{nomencl}
\usepackage[cal=boondox]{mathalfa}
\usepackage{ragged2e}
\usepackage{makecell, boldline}
\definecolor{cerulean}{rgb}{0.0, 0.48, 0.65}
\definecolor{grn}{rgb}{0, .8, .6}
\definecolor{brickred}{rgb}{0.8, 0.25, 0.33}

\makenomenclature

\definecolor{cerulean}{rgb}{0.0, 0.48, 0.65}
\newcommand\norm[1]{\left\lVert#1\right\rVert}

\ifCLASSOPTIONcompsoc
\usepackage[caption=false,font=normalsize,labelfon
t=sf,textfont=sf]{subfig}
\else
\usepackage[caption=false,font=footnotesize]{subfig}
\fi

\begin{document}

\title{An Unsupervised Clustering-Based Short-Term Solar Forecasting Methodology Using Multi-Model Machine Learning Blending}
\author{Cong~Feng,~\IEEEmembership{Student Member,~IEEE,}
        Mingjian~Cui,~\IEEEmembership{Member,~IEEE,}
        Bri-Mathias~Hodge,~\IEEEmembership{Senior~Member,~IEEE,}
        Siyuan~Lu,
        Hendrik F.~Hamann,~\IEEEmembership{Member,~IEEE,}
        and~Jie~Zhang,~\IEEEmembership{Senior~Member,~IEEE}% <-this % stops a space
\thanks{This work was supported by the National Renewable Energy Laboratory under Subcontract No. XHQ-6-62546-01 (under the U.S. Department of Energy Prime Contract No. DE-AC36-08GO28308).}
\thanks{C. Feng, M. Cui, and J. Zhang are with the  Department of Mechanical Engineering,
		The University of Texas at Dallas, Richardson, TX 75080, USA, 
		e-mail: (\{cong.feng1, mingjian.cui, jiezhang\}@utdallas.edu).}% <-this % stops a space
%\thanks{M. Cui is with the  Department of Electrical Engineering,
%	Southern Methodist University, Dallas, TX, 75275, USA, 
%	e-mail: (mingjiancui@smu.edu).}% <-this % stops a space
\thanks{B.-M. Hodge is with the National Renewable Energy Laboratory (NREL),
		Golden, CO 80401, USA, e-mail: (bri.mathias.hodge@nrel.gov).}% <-this % stops a space
\thanks{S. Lu and H. F. Hamann are with IBM TJ Watson Research Center,
		Yorktown Heights, NY 10598, USA, e-mail: (\{lus, hendrikh\}@us.ibm.com).}}% <-this % stops a space	`

%\markboth{IEEE TRANSACTIONS ON SUSTAINABLE ENERGY, 2018}%
\maketitle

\begin{abstract}
Solar forecasting accuracy is affected by weather conditions, and weather awareness forecasting models are expected to improve the performance. However, it may not be available and reliable to classify different forecasting tasks by using only meteorological weather categorization. In this paper, an unsupervised clustering-based (UC-based) solar forecasting methodology is developed for short-term (1-hour-ahead) global horizontal irradiance (GHI) forecasting. This methodology consists of three parts: GHI time series unsupervised clustering, pattern recognition, and UC-based forecasting. The daily GHI time series is first clustered by an Optimized Cross-validated ClUsteRing (OCCUR) method, which determines the optimal number of clusters and best clustering results. Then, support vector machine pattern recognition (SVM-PR) is adopted to recognize the category of a certain day using the first few hours’ data in the forecasting stage. GHI forecasts are generated by the most suitable models in different clusters, which are built by a two-layer Machine learning based Multi-Model (M3) forecasting framework. The developed UC-based methodology is validated by using 1-year of data with six solar features. Numerical results show that (i) UC-based models outperform non-UC (all-in-one) models with the same M3 architecture by approximately 20\%; (ii) M3-based models also outperform the single-algorithm machine learning (SAML) models by approximately 20\%.
\end{abstract}


\IEEEpeerreviewmaketitle

\section*{Nomenclature}
\addcontentsline{toc}{section}{Nomenclature}
\subsection*{A. Acronyms (Alphabetically)}
\begin{IEEEdescription}[\IEEEusemathlabelsep\IEEEsetlabelwidth{$V_1,V_2,V_3,V4$}]
	\item[AIO] All-in-one group.
	\item[ANN] Artificial neural network.
	\item[AHC, DHC] Agglomerative hierarchical clustering, divisive hierarchical clustering.
	\item[CI, CSI] Cloud index, clear sky index.
	\item[GBM] Gradient boosting machine.
	\item[GHI] Global horizontal irradiance.	
	\item[HA, DA] Hour-ahead, day-ahead.
	\item[ML, M3] Machine learning, machine learning based multi-model.
	\item[OCCUR] Optimal cross-validated clustering.
	\item[PV] Photovoltaic.
	\item[PR, SVM-PR] Support vector machine, pattern recognition, support vector machine pattern recognition.
	\item[RBR, nRBR] Red blue ratio, normalized red blue ratio.
	\item[RF] Random forest.
	\item[SAML] Single-algorithm based machine learning.
	\item[SVM, SVR] Support vector machine, support vector regression.	
	\item[UC] Unsupervised clustering.
	%\item[SOM] Self-orgnized map.
\end{IEEEdescription}
\vspace{-.4cm}
\subsection*{B. Variables, Indices, Parameters, Vectors, Matrices, Sets, and Functions:}
\begin{IEEEdescription}[\IEEEusemathlabelsep\IEEEsetlabelwidth{$V_1,V_2,V_3,V4$}]
	\item[$\mathcal{a}$, $\mathcal{b}$] Indices of machine learning algorithm and kernel.
	\item[$C$] Tradeoff parameter of SVM objective function.		
	\item[$CSI$] Variable clear sky  index.
	\item[$\mathcal{C}$, $\boldsymbol{c}$, $\mathcal{M}$, $\boldsymbol{m}$] Centroid sets, centriods, medoid sets, and medoids of clusters.
	\item[$c$, $a$, $m$, $s$] Group indices of unsupervised clustering, all-in-one, multi-model machine learning, and single-algorithm machine learning  groups.	
	\item[$d$, $d_f$, $d_t$] Dimension indices.
	\item[$d_a$, $d_b$] Average distance between an object and other objects in the same cluster, average distance between an object and objects in the nearest neighboring cluster.
	\item[$DNI$, $DHI$] Direct normal irradiance, direct horizontal irradiance.
	\item[$f_{\mathcal{ab}}(\cdot)$, $\Phi(\cdot)$] First-layer and second-layer algorithm.
	\item[$GHI$, $GHI_{clr}$] Global horizontal irradiance and clear sky global horizontal irradiance variables .
	\item[$\boldsymbol{GHI}$, $\boldsymbol{GHI_{clr}}$] Global horizontal irradiance and clear sky global horizontal irradiance sets.
	\item[$k$, $k'$, $k''$] Cluster indices.
	\item[$K$, $K_{max}$, $K_{opt}$] Total number of clusters, maximum $K$, optimal $K$.
	\item[$l$, $L$, $\mathcal{S}^{(l)}$, $k^{(l)}$] Hierarchical level index, total number of hierarchical levels, subset, and cluster index at hierarchical level $l$.		
	\item[$M_{l,ij}$, $M^{i/j}$] Model with kernel $l$ in group $i$ and $j$, comparison of models in group $i$ and $j$.	
	\item[$n$, $n^{(k)}$,$n_b$] Total number of objects in $\mathcal{S}$, $\mathcal{S}_k$, total number of nearest neighbours.	
	\item[$p$, $q$] Indices of total cluster number, method.
	\item[$r_1$, $r_2$, $r_3$, $\boldsymbol{r}$] Voting indices, vector.	
	\item[$R$, $B$, $nRBR$] Blue, red, and normalized red blue ratio in the RGB color system.
	\item[$\mathcal{S}$, $\mathcal{S}_k$] Universal set and clustering disjoint partitions in clustering.
	\item[$T$, $RH$, $Pres$] Temperature, relative humidity, air pressure.	
	\item[$V$, $v^{(K)}$] Vote vector, vote to total cluster number $K$.
	\item[$WS$, $WD$] Wind speed, wind direction.
	\item[$\boldsymbol{x}$, $\boldsymbol{x_{ij}}$, $\boldsymbol{x^{(k)}}$] Clustering objects (data vectors), $j$th nearest neighboring object of $\boldsymbol{x_i}$, and vectors belong to cluster $k$.
	\item[$\boldsymbol{x_i^{(p)}}$, $\boldsymbol{x_i}$, $\boldsymbol{X}$] Input vectors of pattern recognition model and forecasting models, input dataset.
	\item[$\tilde{y}$, $\hat{y}$, $y_i^{(p)}$] Values of the first-layer forecasts, second-layer final forecasts, and output of the pattern recognition model.
	\item[$\boldsymbol{\tilde{Y}}$, $\boldsymbol{\hat{Y}}$] Vectors of the first-layer forecasts and second-layer final forecasts.
	\item[$\alpha$, $\psi$] Weighted vector, bias constant.
	\item[$\beta$] Connectedness measurement.
	\item[$w_{i,k}$] Membership of $\boldsymbol{x_i}$ in cluster $k$.
	\item[$\norm{\cdot}_2$] Euclidean norm.
	\item[$\kappa(\cdot)$, $\varrho$] Kernel function and kernel parameter of SVM-PR.
	\item[$\xi$, $\xi^*$] Upper and lower bands of the deviations around SVM objective function.	
	\item[$\mu$, $\sigma$, $H$] Mean, standard deviation, and R\'enyi entropy of nRBR.
	\item[$\boldsymbol{\mu}$, $\boldsymbol{\sigma}$, $\boldsymbol{H}$] Sets of mean, standard deviation, and R\'enyi entropy of nRBR.

\end{IEEEdescription}
\vspace{-.4cm}
\subsection*{C. Evaluation Metrics:}
\begin{IEEEdescription}[\IEEEusemathlabelsep\IEEEsetlabelwidth{$V_1,V_2,V_3,V4$}]
	\item[$Conn$] Connectiviey index.
	\item[$Silh$] Silhouette width.
	\item[$Dunn$] Dunn's index.	
	\item[$S_{tv}$, $P_{cs}$, $A_{cc}$] Pattern recognition sensitivity, precision, and accuracy.	
	\item[$nMAE$] Forecasting normalized mean absolute error.	
	\item[$nRMSE$] Forecasting normalized root mean square error.	
	\item[$ImpA$] Improvement of forecasting normalized mean absolute error.	
	\item[$ImpR$] improvement of forecasting normalized root mean square error.		
	
\end{IEEEdescription}

\section{Introduction}
\IEEEPARstart{S}{olar} power is a potential alternative to fossil fuel-generated power due to its sustainability. The global installed photovoltaic (PV) capacity is expected to reach 4,600 GW by 2050, providing approximately 16\% electricity worldwide~\cite{philibert2014technology}. The U.S. has installed 47 GW of PV by 2017, with California having the highest solar penetration~\cite{sawin2010renewables}. However, the variability and uncertainty of PV power pose challenges to economic and reliable power system operations. Therefore, improving solar forecasting (including solar power and solar irradiance forecasting) has drawn attention from power system operators, as a means of reducing the uncertainty associated with solar power output.

A collection of statistical and machine learning (ML) methods have been proposed in the literature for short-term solar forecasting. For example, Shakya \textit{et al.}~\cite{shakya2017solar} developed a 1-day-ahead (1DA) solar irradiance forecasting model based on Markove swiching method, which provided solar forecasting for remote areas. Zhang \textit{et al.}~\cite{zhang2015day} compared radial basis function neural networks, least square SVM, k-nearest neighbor (kNN), and weighted kNNs (WkNNs), and found kNN and WkNNs yielded the most competitive forecasting results. A comprehensive review of these methods can be found in latest review papers~\cite{voyant2017machine,antonanzas2016review,raza2016recent}. Even though the learning ability of ML models has been enhanced notably, it is still challenging to capture the complex input-output relationship with single-algorithm based ML (SAML) methods, especially under different conditions. For example, none of single-algorithm ML models outperformed others under all weather conditions in~\cite{gigoni2018day}. On the other hand, the forecasting performance of ML models is critically influenced by the inputs. Some advanced techniques have been explored recently to enhance ML forecasting by providing informative input features, such as total sky images~\cite{feng2018hourly, feng1short}, satellite images~\cite{jang2016solar}, ground-based sensor measurements~\cite{agoua2017short}, and numerical weather predictions~\cite{andrade2017improving}. Among these information sources, features such as historical forecasting variable, cloud index (CI), and red blue ratio (RBR) features of sky images, ground-based weather measurements, and satellite weather data are among the most informative inputs to the ML models.

Solar features are highly influenced by weather conditions. Therefore, there is no guarantee to get an accurate forecast for different weather conditions from a single model. In order to differentiate time series forecasting into different conditions where disparate models can be applied, two processes are required: clustering and classification. Clustering is a process to distinguish and label the type of each time period in the training data. Classification is to identify the category of a time period at the forecasting stage. Several clustering methods have been reported in the literature to divide forecasting into subtasks. For example,  a combination of self orgnizing map and learning vector quantization was used in~\cite{yang2014weather} to distinguish three predefined weather types. K-means clustering was  applied in~\cite{bae2017hourly} to cluster solar irradiance patterns. A pattern discovery method was adopted in~\cite{sanjari2017probabilistic} to classify different PV system classes. A more comprehensive review of clustering and classification methods in the renewable energy domain can be found in~\cite{perez2016review}. However, several drawbacks exist in these methods: (i) most of the existing work uses pre-defined or meteorologically defined criteria (such as weather condition) in the clustering process, which may not be suitable, available, and reliable for clustering methods; (ii) the number of clusters is not optimized for clustering; (iii) adopted clustering methods are not always reliable for data with specific characteristics.

Pattern recognition (PR) is a kind of classification techniques that identify labels of objects. In solar forecasting, PR has been adopted to recognize to which cluster a forecasting object belongs, therefore a suitable model is selected to perform the forecasting. For example, forecasting errors at the current time was used to identify the current pattern and select the corresponding model in the next forecasting step in~\cite{wu2013prediction}. The temperature difference between the forecast day and the current day was employed to identify the weather type in~\cite{ding2011ann}. A support vector machine (SVM) was used to determine weather types using six extracted solar features in~\cite{wang2015solar}. However, existing methods have multiple limitations: (i) PR  is mainly used in 1 day-ahead (DA) or longer time horizon forecasting, which takes advantage of longer input vectors and therefore is theoretically easier than that for shorter-term time horizons; (ii) some models are required to use indirect variables, such as temperature and clear sky index, to determine the weather pattern; (iii) more advanced algorithms are required to improve the PR accuracy. 

To address the aforementioned limitations, we seek to improve solar forecasting by enhancing solar data clustering, PR, and forecasting learning abilities simultaneously. In what follows, an advanced unsupervised clustering (UC) method only utilizes GHI time series without other indirect variable information. Then, PR identifies the cluster to which a forecasting day belongs with first few hours' data. Lastly, a two-layer ML-based Multi-Model (M3) forecasting framework~\cite{feng2017data} is developed in this paper to reinforce learning ability of the ML models. The main innovations and contributions of this paper include: 
\begin{itemize}
		\item[(i)] Developing a novel Optimized Cross-validated ClUsteRing (OCCUR) method to optimize both the number of clusters and the clustering performance;
		\item [(ii)] Adopting an advanced support vector machine pattern recognition (SVM-PR) method to identify categories of forecasting days with a small number of inputs;
		\item [(iii)] Leveraging the powerful learning ability of a two-layer M3 model at the forecasting stage;
		\item [(iv)] Validating the superiority of the developed UC-M3 based method by confirming the effectiveness of both UC and M3 methods under different conditions that consider both calendar and clustering effects.
\end{itemize}

The remainder of this paper is organized as follows. The Optimized Cross-validated ClUsteRing (OCCUR) method is developed in Section II. Section III describes the SVM-PR, M3, and the overall unsupervised cluster based (UC-based) solar forecasting methodology. Numerical simulations are carried out in Section IV to validate the developed methodology. Section V summarizes the conclusions. 
\section{Optimized Cross-validated ClUsteRing (OCCUR)}
Clustering unlabelled daily GHI time series is an unsupervised learning problem, wherein the inherent structure needs to be deduced. Unsupervised learning is far more challenging than supervised learning because of the lack of foreknowledge of the data. The number of clusters also varies by using different UC algorithms and evaluation metrics. In this paper, an Optimized Cross-validated ClUsteRing (OCCUR) method is proposed to optimize and cross-validate the number of clusters using UC algorithms. This OCCUR method adopts multiple UC algorithms to perform clustering separately. The clustering results are cross-validated using several internal validity indices.

\subsection{Unsupervised Clustering (UC) Algorithms}
Four UC algorithms are used in the OCCUR method, which are: K-means, K-medoids, Agglomerative hierarchical clustering (AHC), and divisive hierarchical clustering (DHC). K-means is a widely used UC algorithm.  Given a dataset $\mathcal{S}=\{\boldsymbol{x_1},...,\boldsymbol{x_n}\}$ of $n$ $d$-dimensional vectors, K-means is a partitional clustering method to construct $K$ disjoint subsets $\mathcal{S}=\{\mathcal{S}_1,...,\mathcal{S}_K\}$, such that $\mathcal{S}_k\neq \varnothing$ ($k=1,2,...,K$), $\mathcal{S}_k\bigcap \mathcal{S}_{k'} = \varnothing$ ($k,k'=1,2,...,K$ and $k\neq k'$), and $\bigcup_{k=1}^K \mathcal{S}_k=\mathcal{S}$~\cite{quilumba2015using}. The main idea of the algorithm is to determine the centroids, $\mathcal{C}={\{\boldsymbol{c_1},...,\boldsymbol{c_K}}\}$, and disjoint subsets $\mathcal{S}$ as follows:
\begin{equation}
\label{cluster1}
	w_{i,k}=\begin{cases}
				1, \quad\boldsymbol{x_i}\in \mathcal{S}_k\\
				0, \quad\boldsymbol{x_i}\notin \mathcal{S}_k
			\end{cases}
\end{equation}

\begin{equation}
	\boldsymbol{c_k}=\frac{\sum_{i=1}^{n}w_{i,k}\boldsymbol{x_i}}{\sum_{i=1}^{n}w_{i,k}}
	\label{centriods}
\end{equation}

\begin{equation}
	\mathcal{S}_k=\{\boldsymbol{x^{(k)}}\}=\{\boldsymbol{x^{(k)}_1},...,\boldsymbol{x^{(k)}_{n^{(k)}}}\}
	\label{subsets}
\end{equation}
where $w_{i,k}$ is the data vector membership of $\boldsymbol{x_i}$ in cluster $k$. For example, $w_{1,1}=1$ means $\boldsymbol{x_1}$ belongs to $ \mathcal{S}_1$ and $w_{1,1}=0$ means $\boldsymbol{x_1}$ does not belong to $ \mathcal{S}_1$. $\boldsymbol{x}^{(k)}$ is the data vector categorized into cluster $k$. The K-means algorithm repeats iterative refinement steps by updating the centroids and subsets based on Eqs.~(\ref{centriods}) and~(\ref{subsets}), until reaching the optima as follows~\cite{quilumba2015using, mets2016two}:
\begin{equation}
	\operatornamewithlimits{argmin}_{\mathcal{S}}\sum_{k=1}^{K}\sum_{i=1}^{n}w_{i,k}\norm{(\boldsymbol{x}_i-\boldsymbol{c}_k)}_2
\end{equation}
where $\norm{\cdot}_2$ is the Euclidean norm, which is used to calculate the distance.

Another partitional UC algorithm adopted is the K-medoids method. Instead of clustering based on the centroids, K-medoids seeks the medoids of clusters. A medoid is the most centrally located actual object (data vector in the regression case) within a cluster, which makes K-medoids more robust than the K-means in some cases. Medoids, $\mathcal{M}={\{\boldsymbol{m_1},...,\boldsymbol{m_K}}\}$, are determined by minimizing the summed distance of a data vector to other vectors within the same cluster~\cite{zhang2017dependency}:
\begin{equation}
	\boldsymbol{m}_k=	\operatornamewithlimits{argmin}_{\boldsymbol{x}\in \mathcal{S}_k }\sum_{i=1}^{n}\sum_{j=1}^{n}w_{i,k}w_{j,k}\norm{(\boldsymbol{x}_i-\boldsymbol{x}_j)}_2
	\label{medoids}
\end{equation}
where $\boldsymbol{m}_k$ is the medoid of the cluster $k$. The objective function of the K-medoids method is modified as:
\begin{equation}
	\operatornamewithlimits{argmin}_{\mathcal{S}}\sum_{k=1}^{K}\sum_{i=1}^{n}w_{i,k}\norm{(\boldsymbol{x}_i-\boldsymbol{m}_k)}_2
\end{equation}
where $\mathcal{S}$ and $\mathcal{M}$ are updated in each iteration until the convergence condition is satisfied.

Agglomerative hierarchical clustering (AHC) is a bottom-up unsupervised hierarchical clustering method. Compared with partitional methods, a predefined cluster number $K$ is not required in AHC~\cite{liu2017hierarchical}. AHC constructs the hierarchy by merging the most similar pairs of lower-level nodes from the bottom to the top. In this paper, the distance of two clusters is calculated by the average linkage method, which is defined as the averaged pairwise distance between data vectors from the two clusters~\cite{dahal2014comprehensive}:
\begin{equation}
	\begin{split}
	&\mathcal{S}^{(l)}\rightarrow\mathcal{S}^{(l+1)}:\\ &\operatornamewithlimits{argmin}_{\mathcal{S}^{(l+1)}}\frac{\sum\limits_{i=1}^{n}\sum\limits_{j=1}^{n}w_{i,k^{(l)}}w_{j,k^{'(l)}}\norm{(\boldsymbol{x}_i-\boldsymbol{x}_j)}_2}{\sum\limits_{i=1}^{n}\sum\limits_{j=1}^{n}w_{i,k^{(l)}}w_{j,k^{'(l)}}},\quad k \neq k'
	\end{split}
	\label{AHC}
\end{equation}
where $\mathcal{S}^{(l)} = \{\mathcal{S}_1^{(l)},...,\mathcal{S}_{K^{(l)}}^{(l)}\}$ is the clustering set at hierarchical level $l$ (1 $\leq l$ $\leq L-1$, where 1 is the bottom level and $L$ is the top level). $w_{i,k^{(l)}}$ is the  membership of $\boldsymbol{x}_i$ in cluster $\mathcal{S}_{k^{(l)}}^{(l)}$. $k$ and $k'$ ensure that the average linkage method is applied to two different clusters at a certain hierarchical level. The clustering result is obtained by cutting the hierarchical dendrogram at a certain height.

Another hierarchical clustering method is divisive hierarchical clustering (DHC)~\cite{aggarwal2013data}, which constructs the hierarchy in a top-down manner. DHC splits a cluster into two subclusters until only singletons are obtained. In the splitting process, the bipartitions are determined by maximizing the between-subcluster dissimilarity:
\begin{equation}
	\begin{split}
	&\mathcal{S}^{(l+1)}\rightarrow\mathcal{S}^{(l)}:\\ &\operatornamewithlimits{argmax}_{\mathcal{S}^{(l)}}\frac{\sum\limits_{i=1}^{n}\sum\limits_{j=1}^{n}w_{i,k^{(l)}}w_{j,k^{'(l)}}\norm{(\boldsymbol{x}_i-\boldsymbol{x}_j)}_2}{\sum\limits_{i=1}^{n}\sum\limits_{j=1}^{n}w_{i,k^{(l)}}w_{j,k^{'(l)}}}, \quad k \neq k'
	\end{split}
	\label{DHC}
\end{equation}
where parameters have the same meaning as Eq.~(\ref{AHC}). The same strategy in AHC is used to obtain the clustering results. The complete enumeration splitting is adopted for the optimum in this paper, which can be found in~\cite{ding2002cluster}. 

\subsection{Clustering Assessment Metric}
Evaluating the clustering correctness is challenging due to the absence of data labels. Satisfactory clustering is expected to have desirable \textit{connectedness} among clustering objects, \textit{cohesion} (also known as compactness or homogeneity) within every cluster, and \textit{separation} between clusters. To assess the clustering performance of the aforementioned UC methods, three internal validity indices are adopted to quantify the clustering performance from different perspectives~\cite{handl2005computational, aghabozorgi2015time, brock2011clvalid, munshi2016photovoltaic}.

Connectivity, $Conn$, measures the connectedness between an object and their nearest neighbors, which is expressed as:
\begin{equation}
\label{conn1}
\beta_{\boldsymbol{x_i},\boldsymbol{x_{ij}}}=\begin{cases}
\frac{1}{j}, \quad\boldsymbol{x_i}, \boldsymbol{x_{ij}}\in \mathcal{S}_k\\
0, \quad\boldsymbol{x_i}\in \mathcal{S}_k,\boldsymbol{x_{ij}}\notin \mathcal{S}_k
\end{cases}
\end{equation}

\begin{equation}
		Conn=\sum_{i=1}^{n}\sum_{j=1}^{n_b}\alpha_{\boldsymbol{x_i},\boldsymbol{x_{ij}}}
\end{equation}
where $\boldsymbol{x_{ij}}$ is the $j$th nearest neighbour of $\boldsymbol{x_i}$. $\beta_{\boldsymbol{x_i},\boldsymbol{x_{ij}}}$ is the connectedness measurement between $\boldsymbol{x_{ij}}$ and $\boldsymbol{x_i}$. $n_b$ is the size of nearest neighboring objects. $k = 1,...,K$ is a subset index. A smaller $Conn$ value indicates better clustering performance ($Conn\in(0,+\infty)$).

Silhouette width, $Silh$, quantifies both clustering cohesion and separation. $Silh$ is the average of the Silhouette coefficients of all objects. Silhouette coefficients are calculated based on the distance between a clustering object and other objects within the same cluster, and the distance between the same object and the objects in the nearest neighboring cluster. It is expressed as:
\begin{equation}
d_a(i)=\frac{\sum_{j=1}^{n}w_{i,k}w_{j,k}\norm{(\boldsymbol{x}_i-\boldsymbol{x}_j)}_2}{\sum_{j=1}^{n}w_{j,k}}
\end{equation}

\begin{equation}
d_b(i)=\frac{\sum_{j=1}^{n}w_{i,k}w_{j,k'}\norm{(\boldsymbol{x}_i-\boldsymbol{x}_j)}_2}{\sum_{j=1}^{n}w_{j,k'}}
\end{equation}

\begin{equation}
Silh=\frac{1}{n}\sum_{i=1}^{n}\frac{d_b(i)-d_a(i)}{\max(d_a(i),d_b(i))}
\end{equation}
where $Silh \in [-1,+1]$. $Silh = +1$ indicates desired clustering, vice versa.

The Dunn's index, $Dunn$, is also able to measure both the cohesion and separation of a clustering result, by a ratio between the minimal inter-cluster distance to the maximal intra-cluster distance.
\begin{equation}
\label{dunn}
Dunn = \frac{\min\left\lbrace\sum\limits_{i=1}^{n}\sum\limits_{j=1}^{n}w_{i,k}w_{j,k'}\norm{(\boldsymbol{x}_i-\boldsymbol{x}_j)}_2\right\rbrace}{\max\left\lbrace\sum\limits_{i=1}^{n}\sum\limits_{j=1}^{n}w_{i,k''}w_{j,k''}\norm{(\boldsymbol{x}_i-\boldsymbol{x}_j)}_2\right\rbrace}
\end{equation}
where $k$, $k'$, and $k''$ ensure the independency of the clusters. $Dunn\in[0, +\infty)$, and a larger $Dunn$ indicates better clustering performance.

\subsection{Cross-validation Process}
\begin{algorithm}
\caption{Optimized Cross-validated ClUsteRing (OCCUR) method}
\label{OCCUR}
%\KwIn{a}
%\KwOut{b}
Initialize voting score vector $\boldsymbol{V}=\{v^{(2)},...,v^{(K_{max})}\}$\\
\For{$p\leftarrow 1$ \KwTo $(K_{max}-1)$}{
	\For{$q\leftarrow 1$ \KwTo $4$}{
	Cluster using the $q$th method from Eqs.~\ref{cluster1} - \ref{DHC}, with $(p+1)$ clusters: $\rightarrow\mathcal{S^{p}_q}$\\
	Assess clustering performance based on $\mathcal{S^q_a}$ by Eqs.~\ref{conn1} - \ref{dunn}: $\rightarrow Conn_{pq}$, $Silh_{pq}$, $Dunn_{pq}$\\
	}
} 
	%	\State Construct the evaluation matrics: $\boldsymbol{Conn}=\{Conn_{pq}\}$, $\boldsymbol{Silh}=\{Silh_{pq}\}$, $\boldsymbol{Dunn}=\{Dunn_{pq}\}$
\For{$q\leftarrow 1$ \KwTo $4$}{
Construct evaluation vectors: $\boldsymbol{Conn_q}=\{Conn_{pq}\}$, $\boldsymbol{Silh_q}=\{Silh_{pq}\}$, $\boldsymbol{Dunn_q}=\{Dunn_{pq}\}$. \\
Initialize dynamic evaluation vectors: $\boldsymbol{Conn'_q}=\boldsymbol{Conn_q}$, $\boldsymbol{Silh'_q}=\boldsymbol{Silh_q}$, $\boldsymbol{Dunn'_q}=\boldsymbol{Dunn_q}$\\
\For {$\mathcal{v}\leftarrow 1$ \KwTo $(K_{max}-1)$}{
 Obtain the voting index by sorting evaluation vectors: $r_1=\operatornamewithlimits{argmin}\limits_{p}\boldsymbol{Conn'_q}$, $r_2=\operatornamewithlimits{argmax}\limits_{p}\boldsymbol{Silh'_q}$, $r_3=\operatornamewithlimits{argmax}\limits_{p}\boldsymbol{Dunn'_q}$\\
 Vote the cluster number based on three evaluation metrics: $v^{(r+1)}=K_{max}-\mathcal{v}+v^{(r+1)}$, $\boldsymbol{r}=\{r_1, r_2, r_3\}$\\
 Update the dynamic evaluation metrics by eliminating $Conn_{r_1q}$, $Silh_{r_2q}$, $Dunn_{r_3q}$\\
}
}
Obtain the optimal cluster number: $K_{opt}=\operatornamewithlimits{argmax}\limits_{K}\boldsymbol{V}$
\end{algorithm}
The developed Optimized Cross-validated ClUsteRing (OCCUR) method optimizes the number of clusters, which is expected to be more accurate and reliable than that determined by any single UC method. The optimal cluster number is determined by cross-validating the clustering performance of several UC methods from various perspectives. It is expected to avoid drawbacks of a single clustering method and find the optimum. The pseudocode of the OCCUR method is illustrated in Algorithm~\ref{OCCUR}. The aforementioned four UC methods are adopted to cluster the time series data into a predefined cluster number $K$ $(K=2,...,K_{max})$, which is evaluated by the three internal metrics mentioned above. The clustering results with smaller $Conn$ and larger $Silh$/$Dunn$ values will receive more votes. The optimal cluster number $K_{max}$ is the $K$ with the most votes. The result from the best model is selected as the final clustering result and used in the following PR and forecasting stage.

\section{Pattern Recognition and Classification-based Forecasting}
\subsection{SVM Pattern Recognition}
\begin{figure}[!ht]
	\centering
	\includegraphics[width =2.4in]{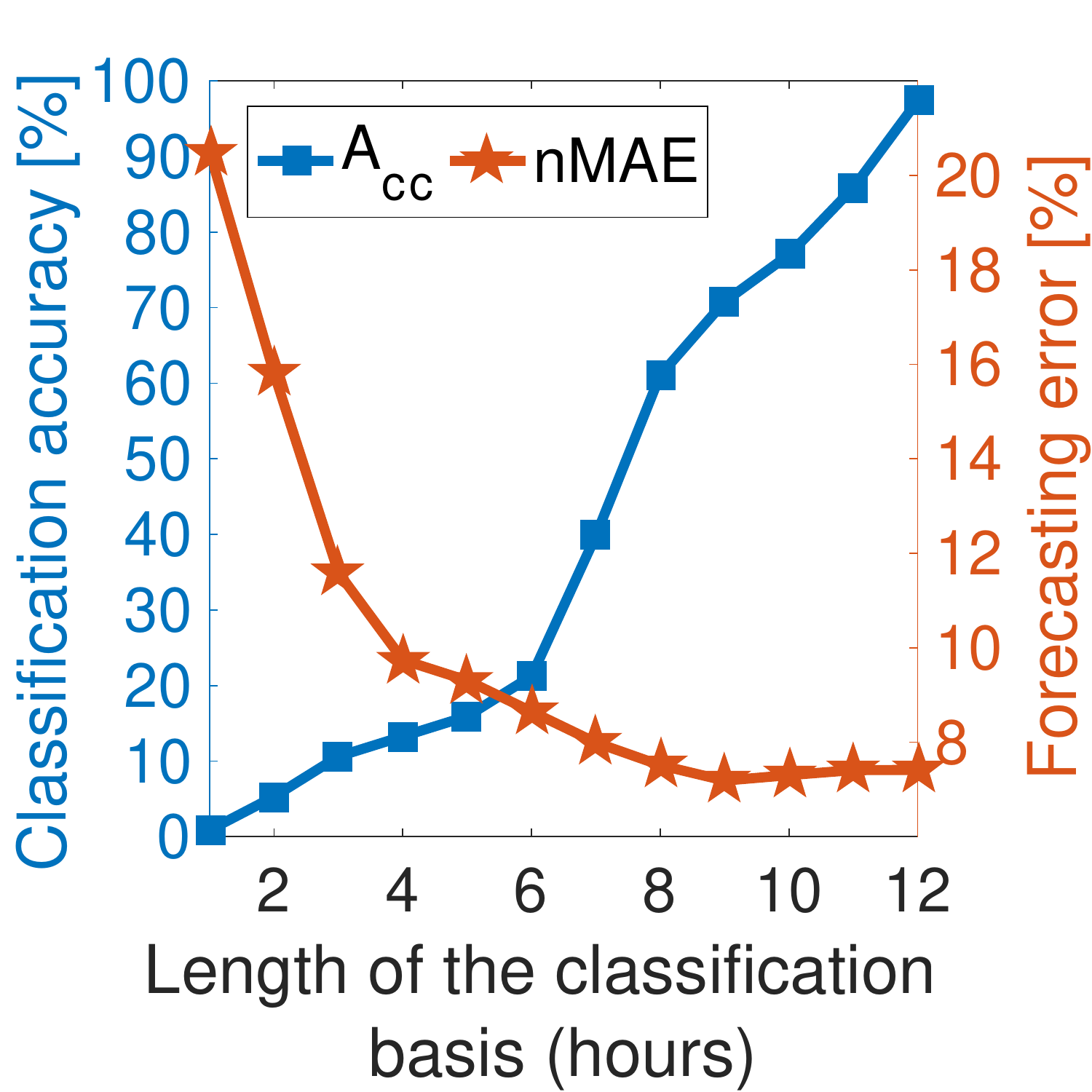}
	\caption{Classification and forecasting accuracy of the direct classification method. The direct classification method and the support vector regression (SVR) method are adopted in the classification and forecasting, respectively~\cite{feng1short}. The overall accuracy ($A_{cc}$) and normalized mean absolute error ($nMAE$) are evaluation metrics to measure the classification and forecasting accuracy, respectively. A larger $A_{cc}$ and a smaller $nMAE$ indicate a better classification and more accurate forecasting, respectively.}
	\label{pr_acc}
\end{figure}

After clustering data into several categories by the OCCUR method, another challenge is to accurately identify groups of the forecasting data. As shown in Fig.~\ref{pr_acc}, the more hours of data used to determine the data cluster label, the more accurate the classification is. The classification accuracy highly affects the forecasting performance. In this case, too many hours (of a day) of data are needed to achieve a satisfying classification and forecasting accuracy, which is impractical in short-term forecasting. Thus, an advanced classification method, SVM PR (SVM-PR), is applied in this paper to identify the data category of a day by using the first few hours' data.

SVM-PR is a classification-based method, which is trained with labeled data and identifies labels in the testing data. To model an SVM classifier, the outputs (weather types) are assumed to take a form of~\cite{wang2015solar}:
\begin{equation} 
y^{(p)}_i=\alpha_i^T\cdot \kappa(\boldsymbol{x^{(p)}_i},\boldsymbol{x'^{(p)}_i})+\psi
\end{equation} 
where $y_i^{(p)}$ and $\boldsymbol{x_i^{(p)}}$ are the output (data cluster label) and $d_x^{(p)}$-dimensional ($d_x^{(p)}=d_f^{(p)}\times d_t^{(p)}$, $d_f^{(p)}$ is the number of features in the PR, and $d_t^{(p)}$ is the number of hours chosen as classification basis) input vector of the SVM-PR model. $\alpha_i$ is a $d_l^{(p)}$-dimensional weighted vector. $\psi$ is the bias constant. $\kappa(\cdot)$ is a kernel function that maps the $d_x^{(p)}$-dimensional input vector into a $d_l^{(p)}$ feature space. A radial basis function (RBF) is selected as the kernel function, expressed as:
\begin{equation} 
\kappa(\boldsymbol{x},\boldsymbol{x'})=e^{-\frac{\|\boldsymbol{x}-\boldsymbol{x'}\|}{2\varrho^2}}
\end{equation}
where $\varrho$ is a kernel parameter. The objective function of the SVM-PR is formulated as:
\begin{equation} 
\rm{min}\quad \frac{1}{2}\|\alpha\|^2+\textit{C}(\sum_{i=1}^{t}(\xi_i+\xi_i^*))
\end{equation}
subject to:
\begin{subequations}
	\begin{align}
	\langle\alpha,\boldsymbol{x_i}\rangle+\psi-y_i\leq\epsilon+\xi_i^*,\ \forall i \\
	y_i-\langle\alpha,\boldsymbol{x_i}\rangle-\psi\leq\epsilon+\xi_i, ,\ \forall i \\
	\xi_i,\ \xi_i^*\geq 0
	\end{align}
\end{subequations}
where $\xi$ and $\xi^*$ are the upper and lower $\epsilon$ bands of deviations around the objective function, respectively. $C$ is a tradeoff parameter. Once the classfier model is trained, the data cluster label can be recognized by an inputs vector $\boldsymbol{x}$ with the same features.

\subsection{Machine Learning-based Multi-Model (M3) Forecasting}
\begin{figure*}[!ht]
	\centering
	\includegraphics[width =7in]{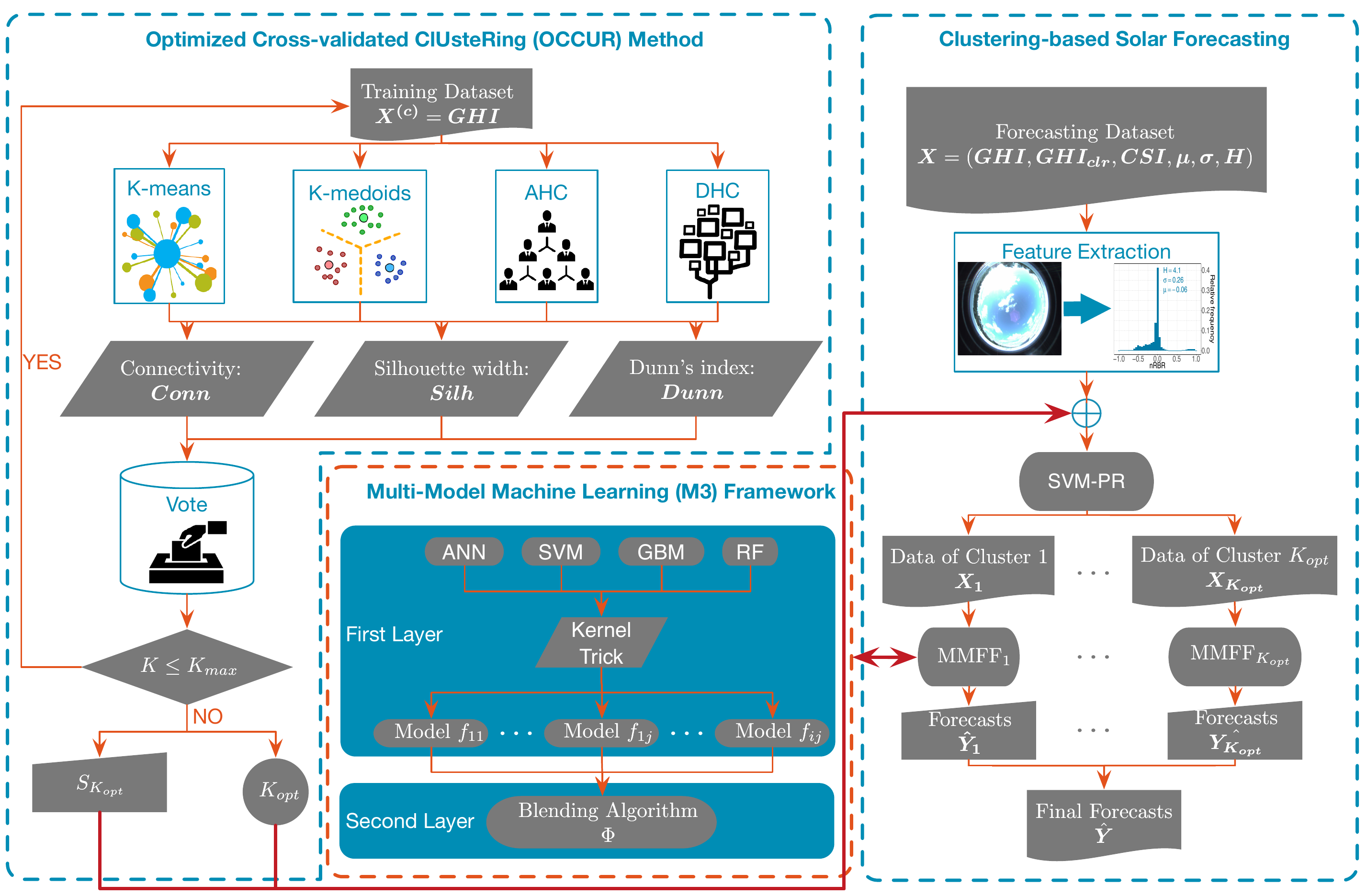}
	\caption{Overall framework of the unsupervised clustering-based short-term solar forecasting methodology.}
	\label{flowchart}
\end{figure*}
 M3 is a two-layer ML based method for short-term forecasting, as shown in the brown box in Fig.~\ref{flowchart}. Multiple ML algorithms with several kernels generate forecasts, $\tilde{\boldsymbol{Y}}$, individually in the first layer. Then the forecasts are blended by a ML algorithm in the second layer, which gives the final forecasts, $\boldsymbol{\hat{Y}}$. ML algorithms used in M3 include artificial neural networks (ANN), SVR, gradient boosting machines (GBM), and random forests (RF). M3 has been shown to perform better than single algorithm ML methods~\cite{feng2017data}. M3 can be expressed as:
\begin{equation} 
\tilde{y}_{i,\mathcal{ab}}=f_{\mathcal{ab}}(\boldsymbol{x}_{i})
\end{equation}

%\vspace{-.5cm}
\begin{equation} 
\hat{y}_i=\Phi(\boldsymbol{\tilde{y}}_{i})
\end{equation}
where $i$ is the data index, $f_{\mathcal{ab}}(\cdot)$ is the model in the first-layer using $\mathcal{a}$th ML algorithm with kernel $\mathcal{b}$, $\tilde{y}_{\mathcal{ab}}$ is the forecast provided by the model $f_{\mathcal{ab}}$, $\boldsymbol{x}_{i}\in \boldsymbol{X}$ is the input vector to the first-layer models, $\boldsymbol{\tilde{y}}=\{\tilde{y}_{\mathcal{ab}}\}$ is the combination of the first-layer forecasts, $\hat{y}_i$ is the final forecast at time $i$, and $\Phi(\cdot)$ is the blending algorithm in the second layer. Note that several blending algorithms can be applied in the second layer, and the best M3 model with a certain blending algorithm in cluster's forecasting is selected to be part of the final forecasting framework (denoted as C$_{opt}$). This training process is evaluated  through a 10-fold cross-validation in the training dataset. More details of M3 can be found in~\cite{feng2017data}.

\subsection{Clustering-based Solar Forecasting}
The UC-M3 solar forecasting integrates OCCUR clustering, SVM-PR, and M3, as shown in Fig.~\ref{flowchart}. The optimal cluster number $K_{opt}$ and the best clustering result $S_{K_{opt}}$ are first determined by OCCUR using the training dataset (only use everyday's GHI). Then, SVR-PR is modeled by labeled $S_{K_{opt}}$, which is adopted to recognize the category of a certain day using the first 4 hours' data (from 7am to 10am, including all the solar features) in the forecasting dataset. M3 is used as the forecasting engine, which is built for each cluster separately. Since most GHIs before 7am are close to zero, which do not provide enough information to build an efficient learning model, GHIs at 7am are forecasted by a 1-day-ahead (1DA) persistence of cloudiness model. This 1DA persistence of cloudiness model assumes a constant clear-sky index within 24 hours, which is expressed by:
	\begin{equation} 
	\begin{aligned}
	GHI_{p}(t+\Delta t)=\frac{GHI(t)}{GHI_{clr}(t)}\times GHI_{clr}(t+\Delta t)
	\end{aligned}
	\end{equation}
	where $GHI_{p}(t+\Delta t)$ means the $GHI$ persistent prediction at time $t+\Delta t$. $GHI$ and $GHI_{clr}$ are GHI measurements and clear-sky GHI values, respectively; $\Delta t$ is the forecasting time horizon of the persistence of cloudiness method, which is 24 in this model. The GHIs at 8am, 9am, and 10am are forecasted by 3 hourly-similarity based M3 models, respectively. For example, the M3$_{8am}$ model is trained by $\{\mathbf{GHI_{7am}}|\mathbf{GHI_{8am}}\}$. More details about hourly-similarity based solar forecasting can be found in~\cite{feng2018hourly, feng1short}. Note that several blending algorithms can be applied in the second layer, and the best M3 model, M3$_{k_\mathcal{a}}$, with a certain blending algorithm in each hour/cluster forecasting is selected to be part of the final forecasting framework. This training process is evaluated  through a 10-fold cross-validation in the training dataset.

\section{Case Study}

\subsection{Data Summary and Feature Extraction}
To obtain well-performing data-driven models, suitable features need to be extracted from different information sources and fed into the models. The features selected in this paper are from three information resources: (i) GHI features: historical GHI ($GHI$), clear sky GHI ($GHI_{clr}$), and clear sky index ($CSI$); (ii) sky imaging features: mean ($\mu$), standard deviation ($\sigma$), and R\'enyi entropy ($H$) of the normalized sky image pixel RBR ($nRBR$) values; and (iii) other meteorological measurements: direct normal irradiance (DNI), direct horizontal irradiance (DHI), temperature (T), relative humidity (RH), pressure (Pres), wind speed (WS), and wind direction (WD).

$GHI_{clr}$ is the GHI value under cloudless conditions, which is generated by a clear-sky model. In this paper, the Ineichen and Perez model \cite{ineichen2002new} is selected as the clear-sky model. $CSI$ is the ratio of $GHI$ and $GHI_{clr}$. The final three features are extracted by sky image processing, and the $nRBR$ of a pixel is calculated by:
\begin{equation} 
nRBR_i = \frac{R_i-B_i}{R_i+B_i}
\end{equation}
where $R_i$ and $B_i$ represent the red and blue values of $i$th sky image pixel in the RGB color system, respectively. The number of pixels in each image is 1392$\times$1040. $nRBR$ is the basis to calculate the three sky imaging features $\mu$, $\sigma$, and $H$. $H$ is the R\'enyi entropy, defined as:
\begin{equation}
H=\frac{1}{1-\gamma}log\left[\sum_{i=1}^{n}(p_i^\gamma)\right]
\end{equation}
where $\gamma=2$ is the order of R\'enyi entropy. $p_i^\gamma$ is the frequency for the $i$th bin (out of 150 evenly spaced bins). These 13 features (i.e., $GHI$, $GHI_{clr}$, $CSI$, $\mu$, $\sigma$, $H$, $DNI$, $DHI$, $T$, $RH$, $Pres$, $WS$, and $WD$) compose the feature space serving as the inputs to the PR model.

A 1-year hourly GHI and sky imaging dataset released by the National Renewable Energy Laboratory (NREL) is adopted in the case study, which was collected at a location in Colorado (latitude = 39.74$^\circ$ North, longitude = 105.18$^\circ$ West, elevation = 1,828.8 m). Since solar feature time series has strong seasonal patterns (strength of seasonality~\cite{Feng2017Characterizing} of GHI is 0.84), the training data is randomly selected from each month,  and the remaining data is used for testing. The ratio of training days to testing days is 3:1. Please note that we assume that by randomly partitioning days into training or testing datasets, the model generality can be better assessed. This data partitioning strategy has been widely used in power system time series forecasting, such as Global Energy Forecasting Competition (GEFCom) 2012~\cite{Hong_2014} and GEFCom 2014~\cite{Hong_2016}. The GHIs at early morning (before 7am) and late night (after 7am) are not included in this paper, since most GHI values during this period are zero.

To validate the developed UC-M3 solar forecasting method, the effectiveness of both UC-based forecasting and M3-based forecasting are proven by comparing two sets of counterparts, which are M3 models vs. single-algorithm ML (SAML) models and UC-based models vs. all-in-one (AIO) models. Therefore, there are totally four groups listed as:
	\begin{itemize}
		\item \textbf{\textit{Group 1}}: UC and M3 (UC-M3) based solar forecasting, which clusters forecasting tasks by OCCUR and adopts M3 as the forecasting engine.
		\item \textbf{\textit{Group 2}}: UC and SAML (UC-SAML) based solar forecasting, which clusters forecasting tasks by OCCUR and adopts normal SAML models as the forecasting engine.
		\item \textbf{\textit{Group 3}}: AIO and M3 (AIO-M3) based solar forecasting, which does not cluster forecasting tasks and adopts M3 as the forecasting engine.
		\item \textbf{\textit{Group 4}}: AIO and SAML (AIO-SAML) based solar forecasting, which does not cluster forecasting tasks and adopts normal SAML models as the forecasting engine.
	\end{itemize}

In each of the above group, several ML algorithms with multiple kernels are adopted to test the generality of the developed UC-based forecasting methodology. Details of these algorithms can be found in~\cite{feng2018hourly}. The experiment is carried out on a laptop with an Intel Core i7 2.6 GHz processor and a 16.0 GB RAM, and the computational time is summarized in Table~\ref{comptime}. The time of forecasting model training varies significantly. UC based models and M3-based models need more time for training than AIO models and SAML models. This is because the UC method has more forecasting models and M3 has two layers.

\begin{table}[!htb]
	\caption{Computational Time (min)}
	\label{comptime}
	\begin{center}  
		\begin{tabular}{lcc}  %left-l,right-r,center-c
			%	\rowcolor{cerulean}&&&&
			%	\\
			\rowcolor{cerulean}\textcolor{white}{\textbf{Process}}&\textcolor{white}{\textbf{Training Time}}&\textcolor{white}{\textbf{PR / Forecasting Time}}\\
			%\rowcolor{cerulean}&&&& \\
			OCCUR&\multicolumn{2}{c}{3.08$\times 10^{-4}$} \\		
			SVR-PR&2.12& 2.12$\times 10^{-4}$\\
			UC-M3 forecasting &8.63&6.81$\times 10^{-2}$\\
			AIO-M3 forecasting&5.78&4.00$\times 10^{-2}$\\
			UC-SAML forecasting&3.66&5.70$\times 10^{-2}$\\
			AIO-SAML forecasting&2.87&2.28$\times 10^{-2}$\\
			\Xhline{1 pt}
		\end{tabular}
	\end{center}
\end{table}

\subsection{OCCUR Clustering Results}
\begin{figure}[!ht]
	\centering
	\includegraphics[width =3.3in]{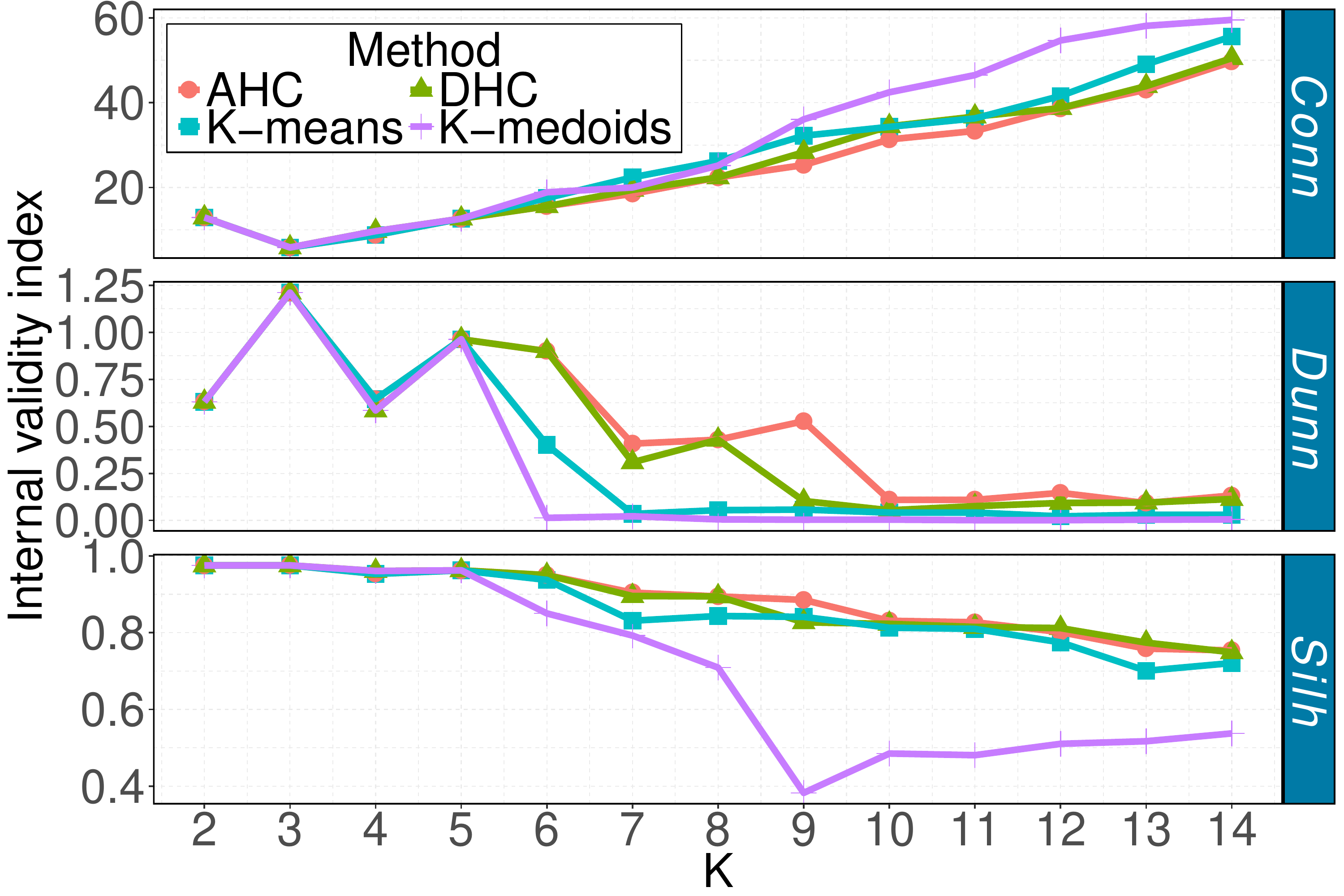}
	\caption{OCCUR clustering results. $K_{max}=14$, $K_{opt}=3$.}
	\label{occur_clst}
\end{figure}
\begin{figure} 
	\centering
	\subfloat[Clustered $GHI$ time series]{%
		\includegraphics[width=0.49\linewidth]{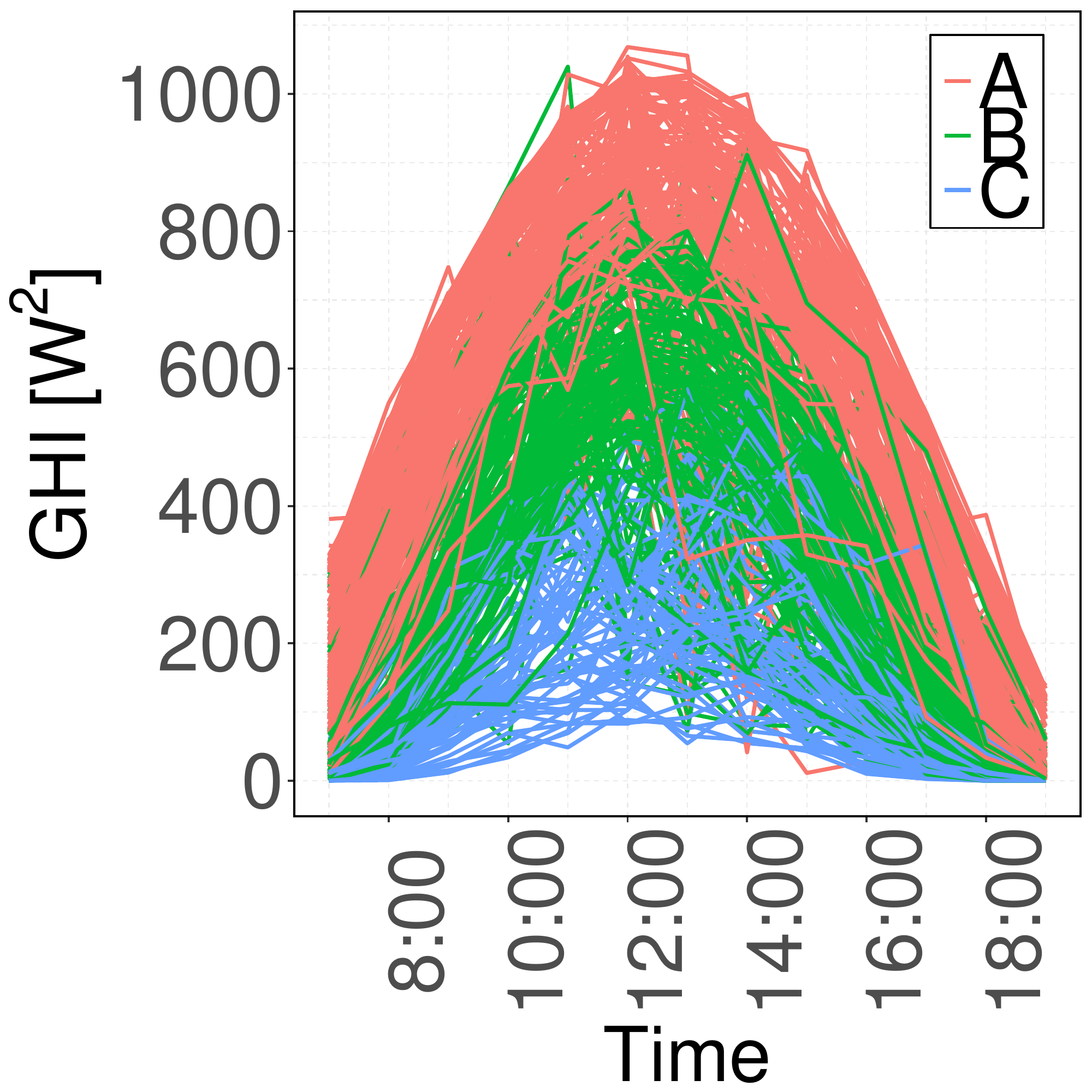}}
	\label{clusterghi1}
	\subfloat[ Clustered $GHI$ statistics]{%
		\includegraphics[width=0.49\linewidth]{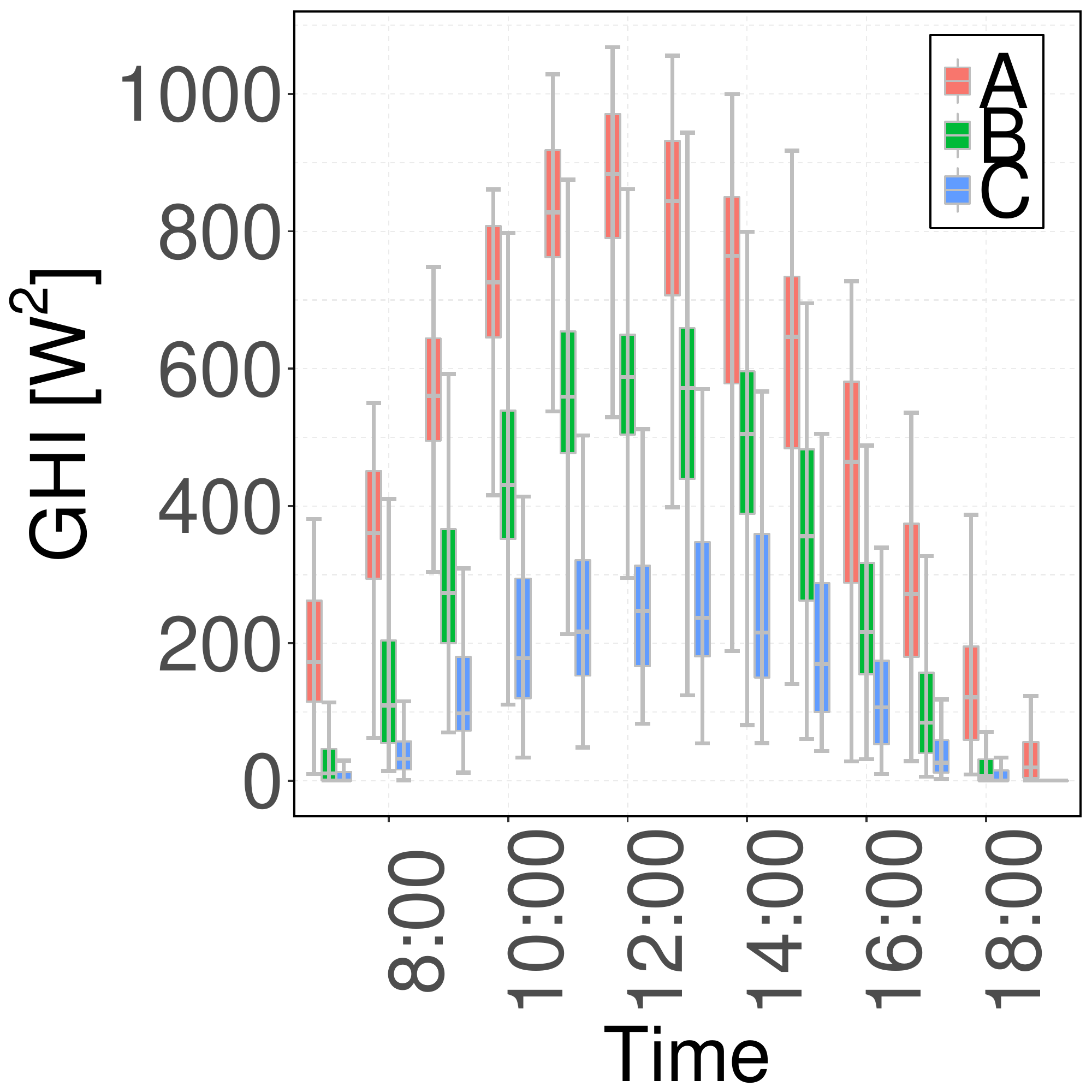}}
	\label{clusterghi2}
	\caption{OCCUR clustering results.}
	\label{occur_result}
\end{figure}

The OCCUR method is carried out to determine the optimal number of clusters first. Fig.~\ref{occur_clst} shows the clustering performance wrapper with $K_{max}=14$. Generally, the connectedness, cohesion, and separation of clustering deteriorate with the increasing number of clusters. When the number of clusters is small ($K \leq 4$), different UC methods illustrate almost equivalent clustering power. With the increasing cluster number, the clustering goodness of different methods becomes distinctive. Fig.~\ref{occur_clst} also shows contrary results evaluated by different internal metrics. For example, the cohesion and separation of the clustering are satisfying but the connectedness is undesirable when $K=2$, compared to $K=3$ and $4$. The cluster number $K=3$ is more suitable than $K=2$ and $4$ by using the $Dunn$ metric; but it's not true when evaluated by $Conn$ and $Silh$. Overall, the optimal number of clusters is $K_{opt}=3$, which is determined by the OCCUR voting process in Algorithm~\ref{OCCUR}. The best UC method when $K=3$ is AHC ($Conn = 5.86$, $Dunn=1.21$, $Silh=0.98$), which is adopted to cluster the training data. The clustered daily $GHI$ time series and corresponding statistics are illustrated in Fig.~\ref{occur_result}. The clustering is evidently layered, which indicates successful clustering.

%\begin{figure}
%	\centering
%	\begin{subfigure}[b]{0.24\textwidth}
%		\centering
%		\includegraphics[width=\textwidth]{Cluster_curves.pdf}
%		\caption{{\small Clustered $GHI$ time series.}}    
%		\label{clusterghi1}
%	\end{subfigure}
%	\hfill
%	\begin{subfigure}[b]{0.24\textwidth}  
%		\centering 
%		\includegraphics[width=\textwidth]{Cluster_stat.pdf}
%		\caption{{\small Clustered $GHI$ statistics.}}    
%		\label{clusterghi2}
%	\end{subfigure}
%	\caption{OCCUR clustering results.}
%	\label{occur_result}
%\end{figure}

\subsection{Pattern Recognition Results}
\begin{figure}[!ht]
	\centering
	\includegraphics[width =2.8in]{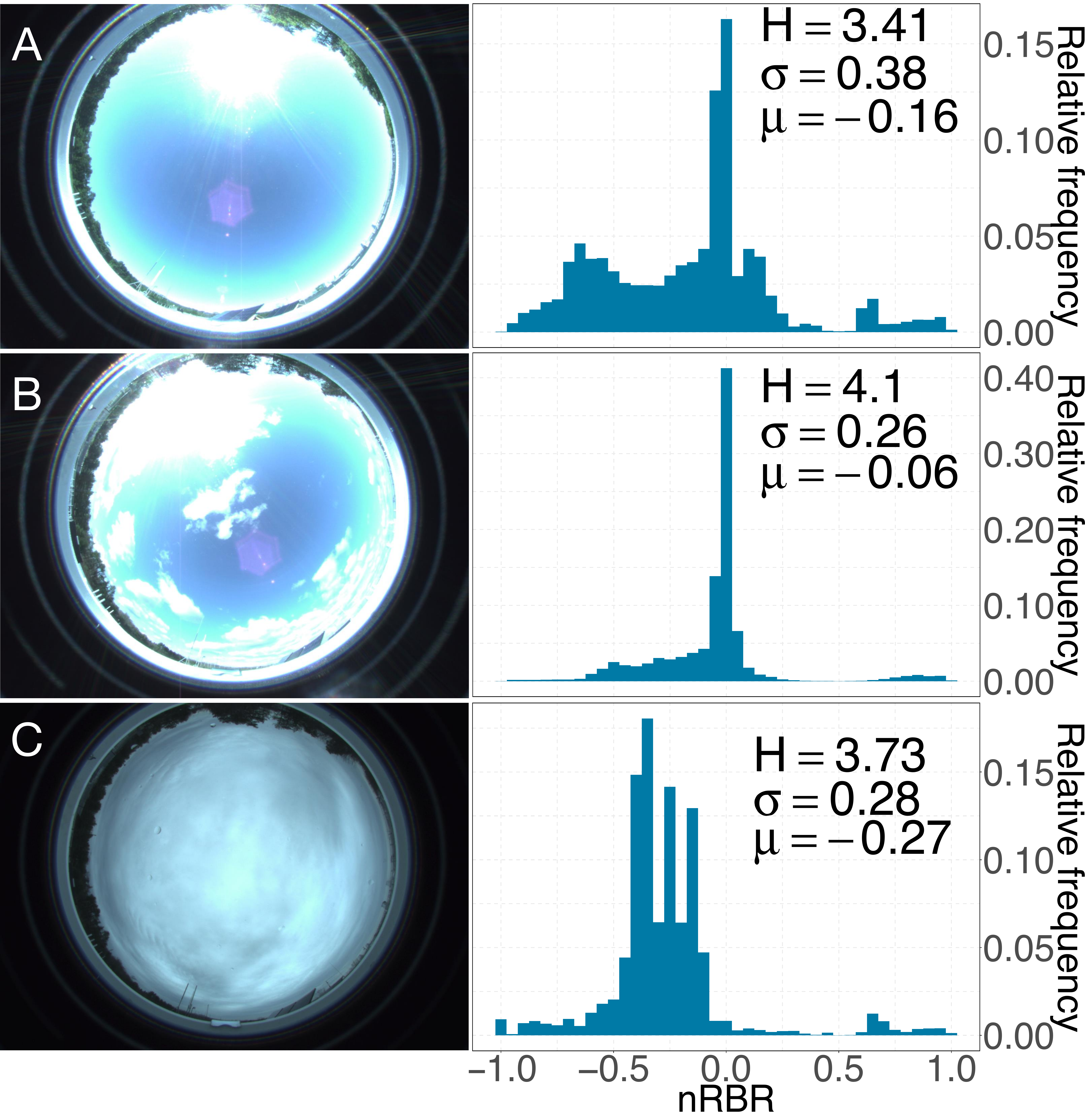}
	\caption{Sky images and corresponding $nRBR$ distributions of 3 clusters.}
	\label{skyimage}
\end{figure}

At the forecasting stage, the category of a certain day is recognized by the first 4 hours' data using the SVM-PR method. All 13 solar features are used in the SVM-PR model. Fig.~\ref{skyimage} shows sky images and their corresponding $nRBR$ distributions in each cluster. Though the clusters are not meteorologically defined, weather features such as cloud cover and irradiance play critical roles in the clustering and PR.

\begin{table}[!htb]
	\caption{Pattern Recognition Results and Evaluation}
	\label{praccuracy}
	\begin{center}  
		\begin{tabular}{llccc}  %left-l,right-r,center-c
			%	\rowcolor{cerulean}&&&&
			%	\\
			\rowcolor{cerulean} && \multicolumn{3}{c}{\textcolor{white}{\textbf{Actual cluster ($k$)}}} \\
			\rowcolor{cerulean}
			\multicolumn{2}{c}{\textcolor{white}{\multirow{-2}{*}{\textbf{Result/evaluation}}}}&\textcolor{white}{\textbf{A}}&\textcolor{white}{\textbf{B}}&\textcolor{white}{\textbf{C}}\\
			%\rowcolor{cerulean}&&&& \\
			\multirow{3}{*}{\textbf{Recognized cluster ($k'$)}}
			&\multicolumn{1}{l|}{\textbf{A}}&36&2&0\\
			&\multicolumn{1}{l|}{\textbf{B}}&3&31&8\\
			&\multicolumn{1}{l|}{\textbf{C}}&1&3&11\\
			\hline
			\multirow{3}{*}{\textbf{PR metrics [\%]}}
			&\multicolumn{1}{l|}{$S_{tv}$}&90.0&86.1&57.9\\
			&\multicolumn{1}{l|}{$P_{cs}$}&94.7&73.8&73.3\\
			&\multicolumn{1}{l|}{$A_{cc}$}&\multicolumn{3}{c}{\textbf{82.1}}\\		
			\Xhline{1 pt}
		\end{tabular}
	\end{center}
\end{table}
Three metrics are used to evaluate the PR results, which are sensitivity ($S_{tv}$), precision ($P_{cs}$), and accuracy ($A_{cc}$). $S_{tv}$ is the proportion of labels that are correctly recognized; $P_{cs}$ is the proportion of recognized labels of a cluster that are correct; and $A_{cc}$ is the proportion of the total number of recognition that are correct. These three metrics are defined as~\cite{wang2015solar}:
\begin{equation}
	S_{tv}=\frac{pr_{kk}}{\sum_{k'=1}^{K_{opt}}pr_{kk'}}
\end{equation}

\begin{equation}
	P_{cs}=\frac{pr_{kk}}{\sum_{k'=1}^{K_{opt}}pr_{k'k}}
\end{equation}

\begin{equation}
	A_{cc}=\frac{pr_{kk}}{\sum_{k=1}^{K_{opt}}\sum_{k'=1}^{K_{opt}}pr_{kk'}}
\end{equation}
where $pr_{kk'}$ represents the objects that belong to cluster $k$ and are recognized to cluster $k'$ ($k$ and $k'$ can be identical). The PR results and performance evaluation are listed in Table~\ref{praccuracy}. Compared to clusters B ($S_{tv}=86.1\%$) and C ($S_{tv}=57.9\%$), objects in cluster A ($S_{tv}=90.0\%$) are recognized more precisely. Most mistakes are made by categorizing the objects into cluster C ($P_{cs}=73.3\%$). By only using the first four hours' data, the overall accuracy is 82.1\%, which is a significant improvement compared to the direct classification method ($A_{cc}$ of the direct classification method using the first four hours' data is only 13\%; to achieve more than 80\% $A_{cc}$, the direct classification method needs more than 11 hours' data, as shown in Fig.~\ref{pr_acc}).

\subsection{Forecasting Results}
In the developed UC-M3 method, once the cluster is recognized by SVM-PR, the most suitable M3 model is selected as the forecasting engine for that specific day (the combination of the suitable models is denoted as C$_{opt}$). Benchmarks include AIO models and SAML models for all clusters in the 4 groups.

\begin{table}[!htb]
	\caption{Overall Forecasting Evaluation}
	\label{frstaccuracy}
	\begin{center}  
		\begin{tabular}{llccccc}  %left-l,right-r,center-c
			%	\rowcolor{cerulean}&&&&&
			%	\\
			%	\rowcolor{cerulean}&&&&&\\
			\rowcolor{cerulean}	&& \multicolumn{2}{c}{\textcolor{white}{\textbf{M3}}}& \multicolumn{2}{c}{\textcolor{white}{\textbf{SAML}}}\\
			\rowcolor{cerulean} \multirow{-2}{*}{\textcolor{white}{\textbf{Group}}}&\multirow{-2}{*}{\textcolor{white}{\textbf{Model}}}&\textcolor{white}{\textbf{$\boldsymbol{nMAE}$}}&\textcolor{white}{\textbf{$\boldsymbol{nRMSE}$}}&\textcolor{white}{\textbf{$\boldsymbol{nMAE}$}}&\textcolor{white}{\textbf{$\boldsymbol{nRMSE}$}}\\
			%   \rowcolor{cerulean}&&&&& \\
			\multirow{13}{*}{\textbf{UC}}
			&\multicolumn{1}{l|}{C$_{opt}$}&{\textcolor{grn}{\textbf{\textit{4.79}}}}&\multicolumn{1}{c|}{{\textcolor{grn}{\textbf{\textit{7.94}}}}}&{\textcolor{grn}{\textbf{\textit{6.37}}}}&{\textcolor{grn}{\textbf{\textit{9.74}}}}\\
			&\multicolumn{1}{l|}{ANN$_{1}$}&5.83&\multicolumn{1}{c|}{9.19}&7.74&11.38\\
			&\multicolumn{1}{l|}{ANN$_{2}$}&5.82&\multicolumn{1}{c|}{9.14}&7.53&11.12\\
			%&\multicolumn{1}{l|}{ANN$_{3,w}$}&6.34&9.68&-13.17&5.39\\
			&\multicolumn{1}{l|}{ANN$_{3}$}&5.86&\multicolumn{1}{c|}{9.19}&7.72&11.42\\
			&\multicolumn{1}{l|}{ANN$_{4}$}&5.71&\multicolumn{1}{c|}{9.11}&8.50&13.73\\
			&\multicolumn{1}{l|}{SVR$_{1}$}&6.86&\multicolumn{1}{c|}{10.53}&8.19&12.15\\
			&\multicolumn{1}{l|}{SVR$_{2}$}&5.70&\multicolumn{1}{c|}{9.66}&{\textbf{6.88}}&\textbf{10.03}\\
			&\multicolumn{1}{l|}{SVR$_{3}$}&5.45&\multicolumn{1}{c|}{\textbf{8.50}}&7.91&11.28\\
			&\multicolumn{1}{l|}{GBM$_{1}$}&5.76&\multicolumn{1}{c|}{8.87}&7.38&10.83\\
			&\multicolumn{1}{l|}{GBM$_{2}$}&5.72&\multicolumn{1}{c|}{8.87}&7.39&10.85\\
			&\multicolumn{1}{l|}{GBM$_{3}$}&\textbf{5.44}&\multicolumn{1}{c|}{8.99}&7.30&11.17\\
			%&\multicolumn{1}{l|}{GBM$_{4}$}&5.71&\multicolumn{1}{c|}{9.11}&\textbf{6.78}&10.37\\
			&\multicolumn{1}{l|}{RF}&5.84&\multicolumn{1}{c|}{9.41}&7.21&10.87\\
			\hline
			\multirow{13}{*}{\textbf{AIO}}
			&\multicolumn{1}{l|}{ANN$_{1}$}&8.04&\multicolumn{1}{c|}{11.58}&10.20&14.63\\
			&\multicolumn{1}{l|}{ANN$_{2}$}&7.61&\multicolumn{1}{c|}{10.97}&10.18&14.42\\
			%&\multicolumn{1}{l|}{ANN$_{3,b}$}&8.06&12.42\\
			&\multicolumn{1}{l|}{ANN$_{3}$}&7.61&\multicolumn{1}{c|}{11.00}&10.25&14.64\\
			&\multicolumn{1}{l|}{ANN$_{4}$}&7.74&\multicolumn{1}{c|}{11.51}&11.55&16.37\\
			&\multicolumn{1}{l|}{SVM$_{1}$}&7.44&\multicolumn{1}{c|}{11.47}&10.51&14.08\\
			&\multicolumn{1}{l|}{SVM$_{2}$}&\textbf{6.40}&\multicolumn{1}{c|}{\textbf{9.73}}&8.39&11.46\\
			&\multicolumn{1}{l|}{SVM$_{3}$}&7.52&\multicolumn{1}{c|}{10.64}&8.59&11.49\\
			&\multicolumn{1}{l|}{GBM$_{1}$}&7.29&\multicolumn{1}{c|}{10.98}&8.53&11.78\\
			&\multicolumn{1}{l|}{GBM$_{2}$}&7.21&\multicolumn{1}{c|}{10.89}&8.48&11.75\\
			&\multicolumn{1}{l|}{GBM$_{3}$}&7.79&\multicolumn{1}{c|}{11.86}&9.49&12.91\\
			%&\multicolumn{1}{l|}{GBM$_{4}$}&7.31&\multicolumn{1}{c|}{10.90}&8.32&12.31\\
			&\multicolumn{1}{l|}{RF}&7.83&\multicolumn{1}{c|}{12.20}&9.40&13.56\\
			&\multicolumn{1}{l|}{P}&7.91&\multicolumn{1}{c|}{11.33}&\textbf{7.91}&\textbf{11.33}\\			
			\Xhline{1.2 pt}
		\end{tabular}
	\end{center}
	\small Note: The units of all the evaluation metrics are \%. The footnotes of models indicate kernel index of the same ML algorithm. `P' represents the 1HA persistence of cloudiness method.%The models with the same model name and numerical subscripts (if any) use the same algorithm. The letter subscript ($c$ or $a$) of the models indicates the group it belongs to. 
\end{table}

\subsubsection{Forecasting accuracy assessment} 
Two commonly used error metrics are used to evaluate forecasting results, which are normalized mean absolute error ($nMAE$) and normalized root mean square error ($nRMSE$)~\cite{feng1short}. The forecasting errors of UC-M3, UC-SAML, AIO-M3, and AIO-SAML groups are listed in Table~\ref{frstaccuracy}. The best UC-M3 (the upper part of Table~\ref{frstaccuracy}) and AIO-M3 models are two $C_{opt}$ models (as highlighted in bold italics). If only a single algorithm is allowed in the M3 second-layer or in a SAML model for different clusters in UC-based forecasting (excluding the $C_{opt}$), SVR$_3$ and GBM$_3$ are the two best UC-M3 models, and SVM$_2$ and GBM$_4$ are the two best UC-SAML models (as highlighted in bold). For the AIO forecasting strategy (the lower part of Table~\ref{frstaccuracy}), SVR$_2$ and 1HA persistence of cloudiness model outperform other M3 models and SAML models, respectively. Compared to other models in the four groups, UC-M3 based $C_{opt}$ model presents the smallest forecasting $nMAE$ and $nRMSE$ values (as highlighted in green).

\subsubsection{Superiorities of UC-based and M3 forecasting}
The superiority of a model over another model can be validated by its forecasting error reduction. Thus, $nMAE$ improvement ($ImpA$) and $nRMSE$ improvement ($ImpR$) are selected in this paper to perform comparisons between UC-based/AIO-based forecasting and M3/SAML forecasting. The ($ImpA$) and ($ImpR$) metrics are defined as:
\begin{equation} 
	ImpA^{j/k} \equiv \frac{nMAE_{M_{l,ij}}-nMAE_{M_{l,ik}}}{nMAE_{M_{l,ij}}}
\end{equation}
\begin{equation} 
	ImpR^{j/k} \equiv \frac{nRMSE_{M_{l,ij}}-nRMSE_{M_{l,ik}}}{nRMSE_{M_{l,ij}}}
\end{equation}
where $M$ is the model name, and $l$ is the kernel index. $i$, $j$, $k$ are group indices, which could be $c$ (UC-based group), $a$ (all-in-one group), $m$ (M3 forecasting group), or $s$ (SAML forecasting group). $ImpA^{j/k}$ and $ImpR^{j/k}$, respectively, are the $nMAE$ and $nRMSE$ improvements of a model in group $j$ compared to the same model in group $k$.

\begin{figure}[!ht]
	\centering
	\includegraphics[width =3.5in]{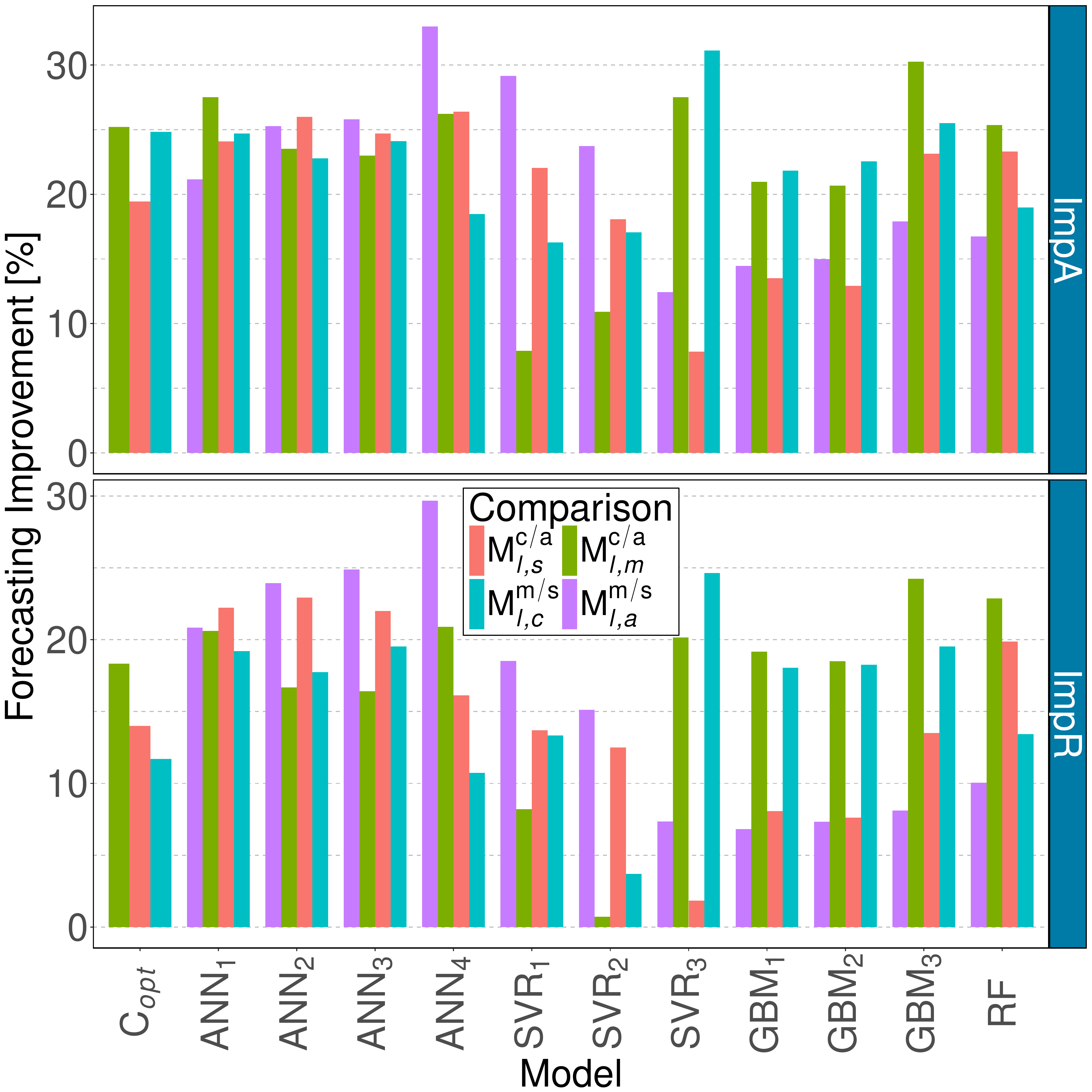}
	\caption{Improvements of UC-based forecasting over AIO-based forecasting and M3 forecasting over SAML forecasting.}
	\label{improv_bar}
\end{figure}

In this paper, the superiority of the developed UC-M3 solar forecasting method is validated by confirming the effectiveness of both UC-based forecasting and M3-based forecasting. Hence, four comparison counterparts are set, which are UC-SAML/AIO-SAML based forecasting (${M_{l,s}^{c/a}}$), UC-M3/AIO-M3 based forecasting (${M_{l,m}^{c/a}}$), UC-M3/UC-SAML based forecasting (${M_{l,c}^{m/s}}$), and M3-AIO/SAML-AIO based forecasting (${M_{l,a}^{m/s}}$). Fig.~\ref{improv_bar} visualizes the above four comparisons, from which several findings are observed. First, both UC and M3 improve the short-term solar forecasting, since all the $ImpA$ and $ImpR$ values are positive. Second, in the same comparison group, the improvements of different models vary distinctively. For example, the $ImpA$ in ${M_{l,m}^{c/a}}$ comparison group ranges from 7.89\% (SVR$_{1,m}$) to 30.25\% (GBM$_{3,m}$). Third, the same model achieves different degrees of improvements combined with different forecasting strategies. For instance, the UC-M3 SVR$_2$ (SVR$_{2,cm}$) model shows only 0.72\% $ImpR$ compared to AIO-M3 SVR$_2$ (SVR$_{2,am}$). However, it reduces 15.13\% $nRMSE$ by using the AIO-M3 SVR$_2$ model (SVR$_{2, am}$) compared with using the AIO-SAML SVR$_2$ model (SVR$_{2, as}$). The average $ImpA^{c/a}$ and $ImpR^{c/a}$ are 21.04\% and 15.51\%; and the average $ImpA^{m/s}$ and $ImpR^{m/s}$ are 21.63\% and 16.36\%, respectively. Therefore, it can be concluded that both UC and M3 have improved the short-term GHI forecasting accuracy significantly.

\subsubsection{Calendar and weather effects}
It is reported in the literature that the forecasting accuracy of power time series, such as solar and load, is influenced by calendar effects~\cite{feng2018short} and weather effects~\cite{gigoni2018day}. To further explore the calendar and weather effects on the developed method, the best model(s) in each group is(are) picked out to make comparisons, which are C$_{opt}$, SVR$_3$, and GBM$_3$ in the UC-M3 group (\{C$_{opt, cm}$, SVR$_{3, cm}$, GBM$_{3, cm}$\}$\in$$M_{l,cm}$), C$_{opt}$ and SVR$_2$ in the UC-SAML group (\{C$_{opt, cs}$, SVR$_{2, cs}$\}$\in$$M_{l,cs}$), SVR$_2$ in the AIO-M3 group (SVR$_{2, am}$$\in$$M_{l,am}$), and 1HA persistence of cloudiness method in the AIO-SAML group (P$_{as}$$\in$$M_{l,as}$).

\begin{figure} 
	\centering
	\subfloat[Forecasting errors by month of the year]{%
		\includegraphics[width=1\linewidth]{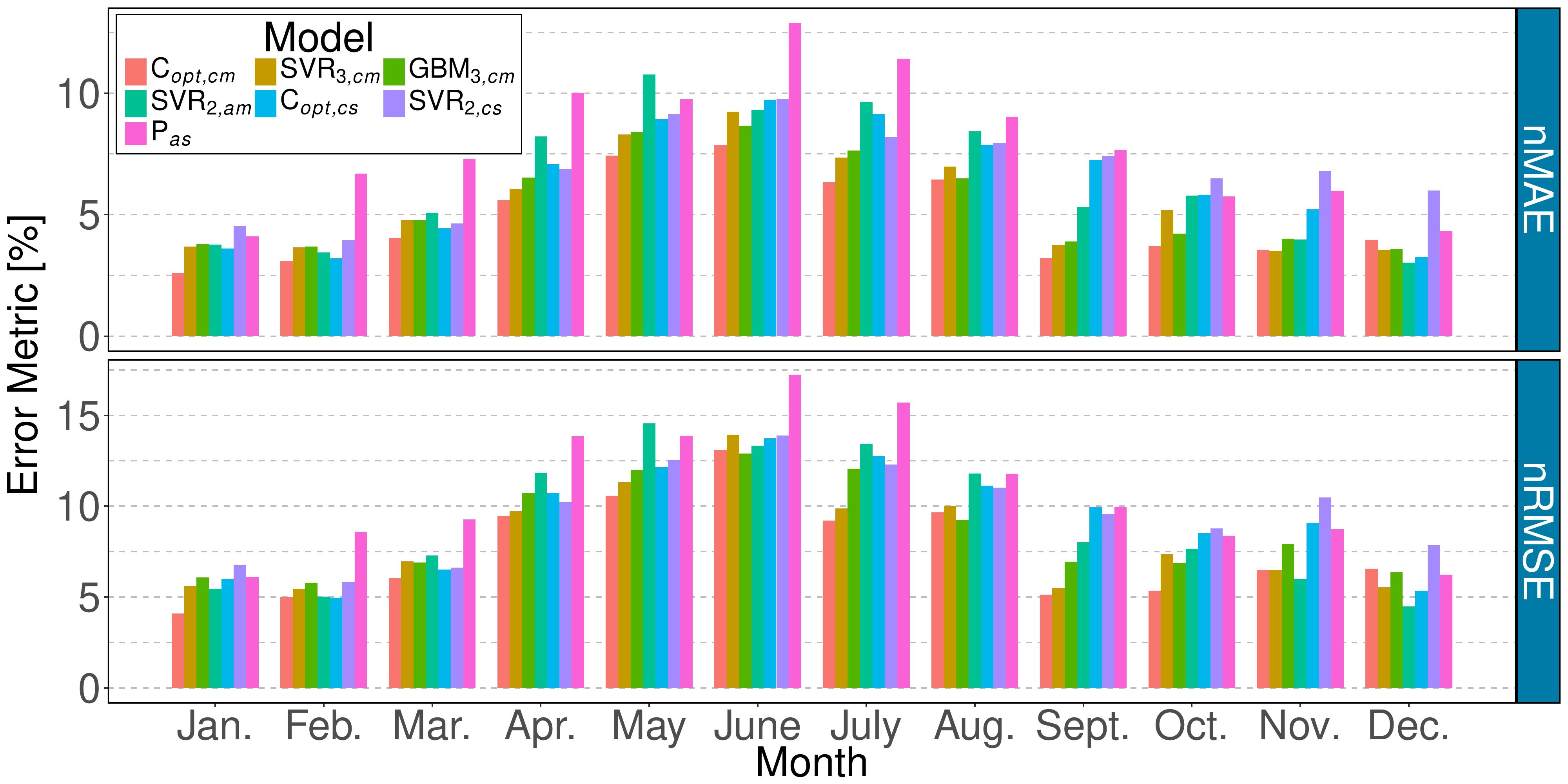}}
		\label{error_month}
	\subfloat[Forecasting errors by hour of the day]{%
		\includegraphics[width=1\linewidth]{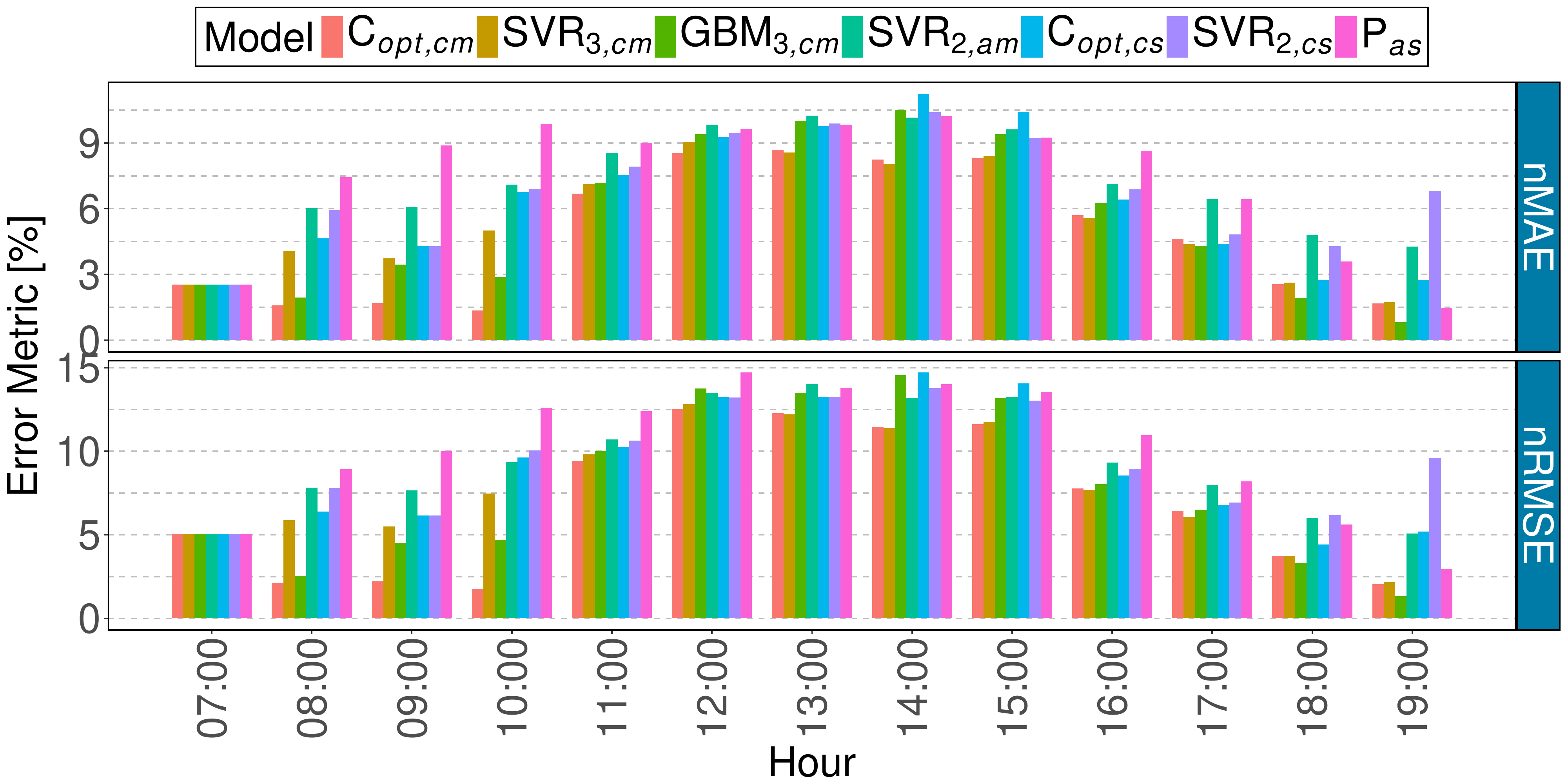}}
		\label{error_hour}
	\caption{Calendar effects on forecasting errors.}
	\label{error_bar1}
\end{figure}
%\begin{figure}[!ht]
%	\centering
%	\begin{subfigure}[b]{.47\textwidth}
%		\centering
%		\includegraphics[width=\textwidth]{error_month.pdf}
%		\caption{{\small Forecasting errors by month.}}    
%		\label{error_month}
%	\end{subfigure}
%	\hfill
%	\begin{subfigure}[b]{.47\textwidth}  
%		\centering 
%		\includegraphics[width=\textwidth]{error_hour.pdf}
%		\caption[]%
%		{{\small Forecasting errors by hour.}}
%		\label{error_hour}
%	\end{subfigure}
%	\caption{Calendar effects on forecasting errors.}
%	\label{error_bar1}
%\end{figure}

Fig.~\ref{error_bar1} presents forecasting errors of the selected 7 models with respect to calendar units (i.e., month of the year and hour of the day). It is observed that forecasting errors show evident daily and yearly patterns due to calendar effects. The forecasting errors are larger in months or hours that have larger GHI values, such as May - Aug. or 11:00 - 15:00. It is also found that the $M_{l,cm}$ and $M_{l,am}$ models show superior performance than those of their counterpart groups (i.e., $M_{l,cs}$ and $M_{l,as}$) in most months and hours. Similarly, the $M_{l,cm}$ and $M_{l,cs}$ models generate smaller forecasting errors than the counterparts in $M_{l,am}$ and $M_{l,as}$ in most months and hours, respectively. Compared to other 6 models, C$_{opt, cm}$ presents better forecasting accuracy in most cases though forecasting error patterns vary due to calendar effects.
%Further analysis is carried out by comparing the model performance in each month and cluster. Four models are selected to compare forecasting errors under different conditions, which are the two best models in each group ({$C_{opt,c}$, P$_a$}$\in$ $M_{l,}$), and the two best MMFF models (based on $nMAE$) with the unique blending algorithm in the two groups (SVM$_{2,c}$ and GBM$_{2,a}$). Compared to all-in-one models, the UC-based models present superior performance in most months, as illustrated in Fig.~\ref{error_month}. Figure~\ref{error_cluster} shows the forecasting errors in three clusters. It is seen that the UC-based models outperform the all-in-one models, while P$_a$ is competitive in cluster-C forecasting. The forecasting improvements of five typical UC-based models, including C$_{opt,c}$ and the other four unique blending algorithm MMFFs, are compared in Fig.~\ref{imp_cluster}. All of the UC-based models enhance the forecasting accuracy in all three clusters. \textit{Thus, overall the UC-based models successfully improve the forecasting accuracy in all of the forecasting subtasks}.
\begin{figure} 
	\centering
	\subfloat[Forecasting errors by cluster]{%
		\includegraphics[width=.49\linewidth]{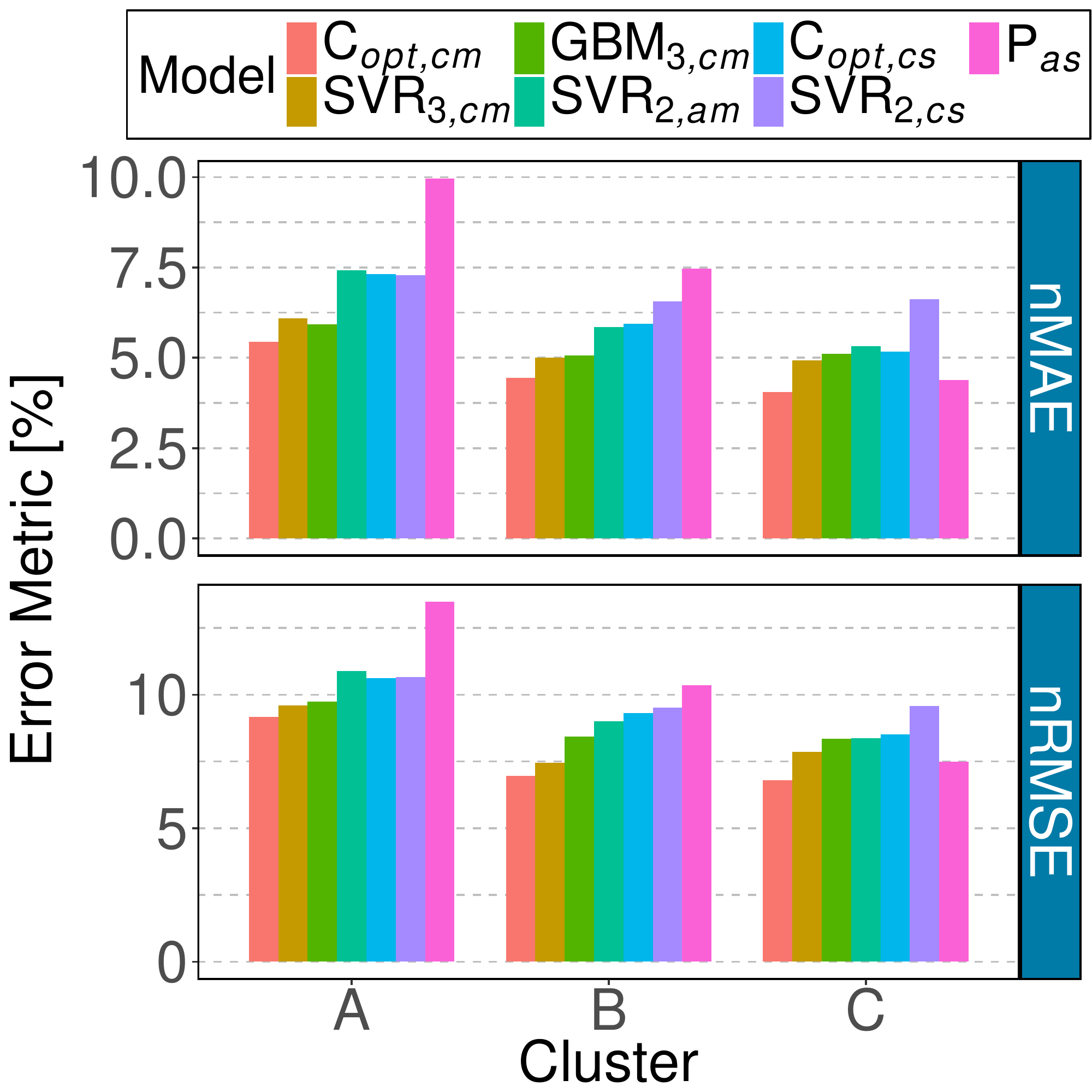}
		\label{error_cluter}}
	\subfloat[Forecasting improvements by cluster]{%
		\includegraphics[width=.49\linewidth]{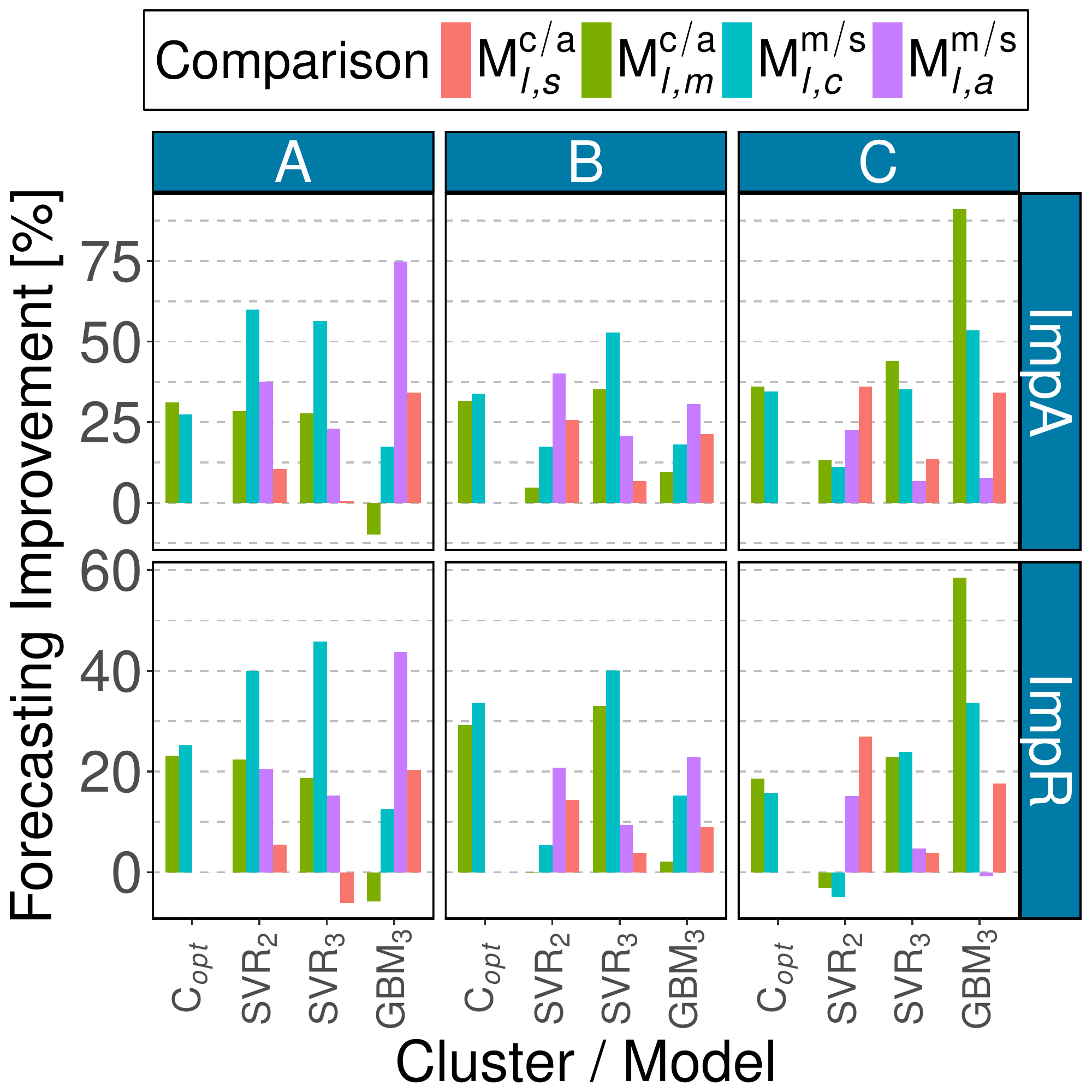}
		\label{errorimp_cluster}}
	\caption{Weather effects on forecasting errors.}
	\label{error_bar2}
\end{figure}
%\begin{figure}[!ht]
%	\centering
%	\begin{subfigure}[b]{.22\textwidth}
%		\centering
%		\includegraphics[width=\textwidth]{error_cluster.pdf}
%		\caption{{\small Forecasting errors by cluster.}}    
%		\label{error_cluter}
%	\end{subfigure}
%	\hfill
%	\begin{subfigure}[b]{.23\textwidth}  
%		\centering 
%		\includegraphics[width=\textwidth]{error_improve_cluster.pdf}
%		\caption[]%
%		{{\small Forecasting improvements by cluster.}}    
%		\label{errorimp_cluster}
%	\end{subfigure}
%	\caption{Weather effects on forecasting errors.}
%	\label{error_bar2}
%\end{figure}

Another way to compare forecasting performance of the developed method is to consider weather conditions. Gigoni \textit{et al.}~\cite{gigoni2018day} evaluated weather effects based on CSI conditions. In this paper, the weather effects on solar forecasting are explored by comparing the best model(s) in each group directly based on 3 clusters. Fig.~\ref{error_bar2} presents forecasting errors and improvements of several best models by cluster. It is observed that models generate smaller errors in cluster C days than in cluster A and cluster B days. This is because cluster A and cluster B have time series with larger GHI values, which lead to larger forecasting variations. Fig.~\ref{error_cluter} also shows that the developed UC-M3 model C$_{opt,cm}$ outperforms other models consistently.  It is found from Fig.~\ref{errorimp_cluster} that both UC and M3 improve the forecasting accuracy significantly in most cases (cluster/models). For example, the M3 model with GBM$_3$ as the blending algorithm in the second layer improves cluster C forecasting accuracy by more than 80\% and 50\% based on $ImpA$ and $ImpR$, respectively. In some cases, such as comparing SVR$_{2,m}^{c/a}$ and comparing SVR$_{2,c}^{m/s}$, the accuracy of SVR$_2$ deteriorates by adopting UC-M3 in cluster C. However, the SVR$_2$ model is significantly improved by UC-M3 in clusters A and B forecasting, which compensates for the deterioration in cluter C. Overall, both UC and M3 have improved solar forecasting in each cluster significantly.

\section{Conclusion}
This paper developed an unsupervised clustering and multi-model machine Learning blending (UC-M3) based methodology to aid short-term global horizontal irradiance (GHI) forecasting. An Optimized Cross-validated ClUsteRing (OCCUR) method was developed to determine the optimal number of clusters and generate the best daily GHI time series clustering. Then, support vector machine pattern recognition (SVM-PR) was utilized to recognize the cluster label of a forecasting day, only using the first four hours' solar data (including sky images, GHI features, and weather information). Finally, UC-M3 based forecasting was carried out by choosing the most suitable M3 model for each clustered forecasting subtask. Case studies based on 1-year of solar data showed that:
	\begin{itemize}
		\item[(1)] The OCCUR method successfully clustered daily GHI time series by using different cross-validated unsupervised clustering methods.
		\item [(2)]SVM-PR enhanced classification accuracy by using limited data within a day (four hours' data in the case study), which made it possible to perform UC-based forecasting.
		\item [(3)]The UC-M3 forecasting method (in conjunction with M3) significantly improved the short-term GHI forecasting accuracy by the effectiveness of both UC and M3 methods.
		\item [(4)]The calendar and weather effects analysis indicated the robust and consistent improvements of the developed UC-M3 method.
	\end{itemize}

%\section*{Acknowledgment}
%This work was supported by the National Renewable Energy Laboratory under Subcontract No. XHQ-6-62546-01 (under the U.S. Department of Energy Prime Contract No. DE-AC36-08GO28308).

\bibliographystyle{IEEEtran}

\bibliography{IEEEfull,solarreference2}

 \newcommand{\noop}[1]{}
\begin{thebibliography}{10}
\providecommand{\url}[1]{#1}
\csname url@samestyle\endcsname
\providecommand{\newblock}{\relax}
\providecommand{\bibinfo}[2]{#2}
\providecommand{\BIBentrySTDinterwordspacing}{\spaceskip=0pt\relax}
\providecommand{\BIBentryALTinterwordstretchfactor}{4}
\providecommand{\BIBentryALTinterwordspacing}{\spaceskip=\fontdimen2\font plus
\BIBentryALTinterwordstretchfactor\fontdimen3\font minus
  \fontdimen4\font\relax}
\providecommand{\BIBforeignlanguage}[2]{{%
\expandafter\ifx\csname l@#1\endcsname\relax
\typeout{** WARNING: IEEEtran.bst: No hyphenation pattern has been}%
\typeout{** loaded for the language `#1'. Using the pattern for}%
\typeout{** the default language instead.}%
\else
\language=\csname l@#1\endcsname
\fi
#2}}
\providecommand{\BIBdecl}{\relax}
\BIBdecl

\bibitem{philibert2014technology}
C.~Philibert, P.~Frankl, C.~Tam, Y.~Abdelilah, H.~Bahar, Q.~Marchais,
  S.~Mueller, U.~Remme, M.~Waldron, and H.~Wiesner, ``Technology roadmap: solar
  photovoltaic energy,'' Internacional Energy Agency(IEA), Tech. Rep., 2014.

\bibitem{sawin2010renewables}
J.~L. Sawin, E.~Martinot, D.~Barnes, A.~McCrone, J.~Roussell, R.~Sims,
  V.~Sonntag-O'Brien, R.~Adib, J.~Skeen, E.~Musolino \emph{et~al.},
  ``Renewables 2017-{G}lobal status report,'' REN21, Tech. Rep., 2017.

\bibitem{shakya2017solar}
A.~Shakya, S.~Michael, C.~Saunders, D.~Armstrong, P.~Pandey, S.~Chalise, and
  R.~Tonkoski, ``Solar irradiance forecasting in remote microgrids using markov
  switching model,'' \emph{IEEE Trans. on Sustain. Energy}, vol.~8, no.~3, pp.
  895--905, 2017.

\bibitem{zhang2015day}
Y.~Zhang, M.~Beaudin, R.~Taheri, H.~Zareipour, and D.~Wood, ``Day-ahead power
  output forecasting for small-scale solar photovoltaic electricity
  generators,'' \emph{IEEE Trans. on Smart Grid}, vol.~6, no.~5, pp.
  2253--2262, 2015.

\bibitem{voyant2017machine}
C.~Voyant, G.~Notton, S.~Kalogirou, M.-L. Nivet, C.~Paoli, F.~Motte, and
  A.~Fouilloy, ``Machine learning methods for solar radiation forecasting: A
  review,'' \emph{Renew. Energy}, vol. 105, pp. 569--582, 2017.

\bibitem{antonanzas2016review}
J.~Antonanzas, N.~Osorio, R.~Escobar, R.~Urraca, F.~Martinez-de Pison, and
  F.~Antonanzas-Torres, ``Review of photovoltaic power forecasting,''
  \emph{Sol. Energy}, vol. 136, pp. 78--111, 2016.

\bibitem{raza2016recent}
M.~Q. Raza, M.~Nadarajah, and C.~Ekanayake, ``On recent advances in {PV} output
  power forecast,'' \emph{Sol. Energy}, vol. 136, pp. 125--144, 2016.

\bibitem{gigoni2018day}
L.~Gigoni, A.~Betti, E.~Crisostomi, A.~Franco, M.~Tucci, F.~Bizzarri, and
  D.~Mucci, ``Day-ahead hourly forecasting of power generation from
  photovoltaic plants,'' \emph{IEEE Trans. on Sustain. Energy}, vol.~9, no.~2,
  pp. 831--842, 2018.

\bibitem{feng2018hourly}
C.~Feng and J.~Zhang, ``Hourly-similarity based solar forecasting using
  multi-model machine learning blending,'' \emph{arXiv preprint
  arXiv:1803.03623}, 2018.

\bibitem{feng1short}
C.~Feng, M.~Cui, M.~Lee, J.~Zhang, B.-M. Hodge, S.~Lu, and H.~F. Hamann,
  ``Short-term global horizontal irradiance forecasting based on sky imaging
  and pattern recognition,'' in \emph{Proc. IEEE Power Energy Soc. Gen.
  Meeting}.\hskip 1em plus 0.5em minus 0.4em\relax Chicago, IL, USA, 2017.

\bibitem{jang2016solar}
H.~S. Jang, K.~Y. Bae, H.-S. Park, and D.~K. Sung, ``Solar power prediction
  based on satellite images and support vector machine,'' \emph{IEEE Trans.
  Sustain. Energy}, vol.~7, no.~3, pp. 1255--1263, 2016.

\bibitem{agoua2017short}
X.~G. Agoua, R.~Girard, and G.~Kariniotakis, ``Short-term spatio-temporal
  forecasting of photovoltaic power production,'' \emph{IEEE Trans. Sustain.
  Energy}, 2017.

\bibitem{andrade2017improving}
J.~R. Andrade and R.~J. Bessa, ``Improving renewable energy forecasting with a
  grid of numerical weather predictions,'' \emph{IEEE Trans. Sustain. Energy},
  2017.

\bibitem{yang2014weather}
H.-T. Yang, C.-M. Huang, Y.-C. Huang, and Y.-S. Pai, ``A weather-based hybrid
  method for 1-day ahead hourly forecasting of {PV} power output,'' \emph{IEEE
  Trans. Sustain. Energy}, vol.~5, no.~3, pp. 917--926, 2014.

\bibitem{bae2017hourly}
K.~Y. Bae, H.~S. Jang, and D.~K. Sung, ``Hourly solar irradiance prediction
  based on support vector machine and its error analysis,'' \emph{IEEE Trans.
  Power Syst.}, vol.~32, no.~2, pp. 935--945, 2017.

\bibitem{sanjari2017probabilistic}
M.~J. Sanjari and H.~Gooi, ``Probabilistic forecast of {PV} power generation
  based on higher order markov chain,'' \emph{IEEE Trans. Power Syst.},
  vol.~32, no.~4, pp. 2942--2952, 2017.

\bibitem{perez2016review}
M.~P{\'e}rez-Ortiz, S.~Jim{\'e}nez-Fern{\'a}ndez, P.~A. Guti{\'e}rrez,
  E.~Alexandre, C.~Herv{\'a}s-Mart{\'\i}nez, and S.~Salcedo-Sanz, ``A review of
  classification problems and algorithms in renewable energy applications,''
  \emph{Energies}, vol.~9, no.~8, p. 607, 2016.

\bibitem{wu2013prediction}
J.~Wu and C.~K. Chan, ``Prediction of hourly solar radiation with multi-model
  framework,'' \emph{Energy conversion and management}, vol.~76, pp. 347--355,
  2013.

\bibitem{ding2011ann}
M.~Ding, L.~Wang, and R.~Bi, ``An {ANN}-based approach for forecasting the
  power output of photovoltaic system,'' \emph{Procedia Environmental
  Sciences}, vol.~11, pp. 1308--1315, 2011.

\bibitem{wang2015solar}
F.~Wang, Z.~Zhen, Z.~Mi, H.~Sun, S.~Su, and G.~Yang, ``Solar irradiance feature
  extraction and support vector machines based weather status pattern
  recognition model for short-term photovoltaic power forecasting,''
  \emph{Energy and Buildings}, vol.~86, pp. 427--438, 2015.

\bibitem{feng2017data}
C.~Feng, M.~Cui, B.-M. Hodge, and J.~Zhang, ``A data-driven multi-model
  methodology with deep feature selection for short-term wind forecasting,''
  \emph{Appl. Energy}, vol. 190, pp. 1245--1257, 2017.

\bibitem{quilumba2015using}
F.~L. Quilumba, W.-J. Lee, H.~Huang, D.~Y. Wang, and R.~L. Szabados, ``Using
  smart meter data to improve the accuracy of intraday load forecasting
  considering customer behavior similarities,'' \emph{IEEE Trans. Smart Grid},
  vol.~6, no.~2, pp. 911--918, 2015.

\bibitem{mets2016two}
K.~Mets, F.~Depuydt, and C.~Develder, ``Two-stage load pattern clustering using
  fast wavelet transformation,'' \emph{IEEE Trans. on Smart Grid}, vol.~7,
  no.~5, pp. 2250--2259, 2016.

\bibitem{zhang2017dependency}
K.~Zhang, H.~Zhu, and S.~Guo, ``Dependency analysis and improved parameter
  estimation for dynamic composite load modeling,'' \emph{IEEE Trans. on Power
  Syst.}, vol.~32, no.~4, pp. 3287--3297, 2017.

\bibitem{liu2017hierarchical}
Y.~Liu, R.~Sioshansi, and A.~J. Conejo, ``Hierarchical clustering to find
  representative operating periods for capacity-expansion modeling,''
  \emph{IEEE Trans. Power Syst.}, 2017.

\bibitem{dahal2014comprehensive}
O.~P. Dahal, S.~M. Brahma, and H.~Cao, ``Comprehensive clustering of
  disturbance events recorded by phasor measurement units,'' \emph{IEEE Trans.
  on Power Del.}, vol.~29, no.~3, pp. 1390--1397, 2014.

\bibitem{aggarwal2013data}
C.~C. Aggarwal and C.~K. Reddy, \emph{Data clustering: algorithms and
  applications}.\hskip 1em plus 0.5em minus 0.4em\relax CRC press, 2013.

\bibitem{ding2002cluster}
C.~Ding and X.~He, ``Cluster merging and splitting in hierarchical clustering
  algorithms,'' in \emph{Data Mining, 2002. ICDM 2003. Proceedings. 2002 IEEE
  International Conference on}.\hskip 1em plus 0.5em minus 0.4em\relax IEEE,
  2002, pp. 139--146.

\bibitem{handl2005computational}
J.~Handl, J.~Knowles, and D.~B. Kell, ``Computational cluster validation in
  post-genomic data analysis,'' \emph{Bioinformatics}, vol.~21, no.~15, pp.
  3201--3212, 2005.

\bibitem{aghabozorgi2015time}
S.~Aghabozorgi, A.~S. Shirkhorshidi, and T.~Y. Wah, ``Time-series
  clustering--{A} decade review,'' \emph{Information Systems}, vol.~53, pp.
  16--38, 2015.

\bibitem{brock2011clvalid}
G.~Brock, V.~Pihur, S.~Datta, S.~Datta \emph{et~al.}, ``clvalid, an {R} package
  for cluster validation,'' \emph{Journal of Statistical Software}, 2011.

\bibitem{munshi2016photovoltaic}
A.~A. Munshi and A.-R.~M. Yasser, ``Photovoltaic power pattern clustering based
  on conventional and swarm clustering methods,'' \emph{Sol. Energy}, vol. 124,
  pp. 39--56, 2016.

\bibitem{ineichen2002new}
P.~Ineichen and R.~Perez, ``A new airmass independent formulation for the linke
  turbidity coefficient,'' \emph{Sol. Energy}, vol.~73, no.~3, pp. 151--157,
  2002.

\bibitem{Feng2017Characterizing}
C.~Feng, E.~K. Chartan, B.-M. Hodge, and J.~Zhang, ``Characterizing time series
  data diversity for wind forecasting,'' in \emph{Proc. Big Data Computing
  Applications and Technologies (BDCAT), 2017 IEEE/ACM 4th International
  Conference on}.\hskip 1em plus 0.5em minus 0.4em\relax IEEE, 2017.

\bibitem{Hong_2014}
T.~Hong, P.~Pinson, and S.~Fan, ``Global energy forecasting competition 2012,''
  \emph{International Journal of Forecasting}, vol.~30, no.~2, pp. 357--363,
  2014.

\bibitem{Hong_2016}
T.~Hong, P.~Pinson, S.~Fan, H.~Zareipour, A.~Troccoli, and R.~J. Hyndman,
  ``Probabilistic energy forecasting: Global energy forecasting competition
  2014 and beyond,'' \emph{International Journal of Forecasting}, vol.~32,
  no.~3, pp. 896--913, 2016.

\bibitem{feng2018short}
C.~Feng and J.~Zhang, ``Short-term load forecasting with different aggregration
  strategies,'' in \emph{ASME 2018 International Design Engineering Technical
  Conferences and Computers and Information in Engineering Conference}.\hskip
  1em plus 0.5em minus 0.4em\relax American Society of Mechanical Engineers,
  2018.

\end{thebibliography}


%\newpage


\end{document}